\documentclass[11pt, a4paper, logo]{dmml}

\pdftrailerid{redacted}

\usepackage[authoryear, sort&compress, round]{natbib}

\usepackage{times}
\usepackage{latexsym}

\usepackage[T1]{fontenc}

\usepackage[utf8]{inputenc}

\usepackage{microtype}

\usepackage{fix-cm}  

\usepackage{graphicx}
\usepackage{xcolor}
\usepackage{subfig}
\usepackage{multirow}
\usepackage{listings}
\usepackage{wrapfig}
\usepackage[most]{tcolorbox}
\newtheorem{theorem}{Theorem}[section]
\newtheorem{definition}{Definition}[section]
\newtheorem{proposition}{Proposition}[section]
\newtheorem{lemma}{Lemma}[section]

\def\methodName{\textsc{DataAlchemy}}

\title{Is Chain-of-Thought Reasoning of LLMs a Mirage? A Data Distribution Lens}

\correspondingauthor{ \{czhao93, ztan36, pingchua, daweili5, bjiang14, yancheng.wang, yingzhen.yang, huanliu\}@asu.edu}

\renewcommand{\today}{}

\author[1]{Chengshuai Zhao}
\author[1]{Zhen Tan}
\author[1]{Pingchuan Ma}
\author[1]{Dawei Li}
\author[1]{Bohan Jiang}
\author[1]{Yancheng Wang}
\author[1]{Yingzhen Yang}
\author[1]{Huan Liu}

\affil[1]{Arizona State University, USA}

\begin{abstract}
Chain-of-Thought (CoT) prompting has been shown to be effective in eliciting structured reasoning (i.e., CoT reasoning) from large language models (LLMs). Regardless of its popularity, recent studies expose its failures in some reasoning tasks, raising fundamental questions about the nature of CoT reasoning. In this work, we propose a data distribution lens to understand when and why CoT reasoning succeeds or fails. We hypothesize that CoT reasoning reflects a structured inductive bias learned from in-distribution data, enabling models to conditionally generate reasoning trajectories that approximate those observed during training. As such, the effectiveness of CoT reasoning is fundamentally governed by the nature and degree of distribution discrepancy between training data and test queries. Guided by this lens, we dissect CoT reasoning via three dimensions: \textit{task}, \textit{length}, and \textit{format}. To test the hypothesis, we introduce \methodName{}, an abstract and fully controllable environment that trains LLMs from scratch and systematically probes them under various distribution conditions. Through rigorous controlled experiments, we reveal that CoT reasoning is a brittle mirage when it is pushed beyond training distributions, emphasizing the ongoing challenge of achieving genuine and generalizable reasoning. Our code is available at GitHub:~\href{https://github.com/ChengshuaiZhao0/DataAlchemy}{https://github.com/ChengshuaiZhao0/DataAlchemy}.

\end{abstract}

\begin{document}
\maketitle

\section{Introduction}

\begin{wrapfigure}{!th}{0.5\textwidth}
    \centering
    \includegraphics[width=0.8\linewidth]{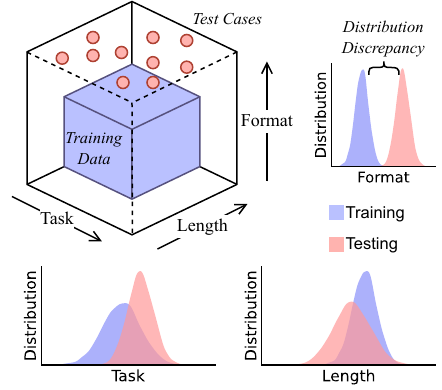}
    \caption{The data perspective lens. CoT reasoning's effectiveness is fundamentally bounded by the degree of distribution discrepancy introduced by \textit{task}, \textit{length}, and \textit{format} between the training data and the test queries.}
    \label{fig:illustration}
\end{wrapfigure}

Chain-of-Thought (CoT) prompting~\citep{wei2022chain} has emerged as a prominent method for eliciting structured reasoning from LLMs (a.k.a., CoT reasoning). By appending a simple cue such as ``Let's think step by step'', LLMs decompose complex problems into intermediate steps, producing outputs that resemble human-like reasoning. It has been shown to be effective in tasks requiring logical inference~\citep{xu2024faithful}, mathematical problem solving~\citep{imani2023mathprompter}, and commonsense reasoning~\citep{wei2022chain}. The empirical success led to CoT reasoning being seen as a promising direction towards artificial general intelligence.

However, some pioneering studies have revealed failures that challenge this optimistic view~\citep{mirzadehgsm}. \citet{stechly2024chain} demonstrate that LLMs fail to generalize in planning tasks, revealing a deficiency in true algorithmic reasoning. \citet{shojaee2025illusion} find that reasoning models experience an accuracy collapse in puzzle-solving once task complexity exceeds a critical threshold. \citet{sun2025omega} demonstrate that LLMs struggle to solve complex mathematical reasoning problems, failing to integrate or adapt learned skills to novel or creative tasks.

Considering the above \emph{opposing} opinions, there is clearly no \textit{indisputable lens to understand why and when CoT reasoning succeeds or fails}. Current evaluation approaches have intrinsic limitations that prevent them from answering the Why and When questions: (\textit{i}) Narrowly defined settings. Existing frameworks focus on specific tasks and evaluate them using specific LLMs, thereby overlooking common structural patterns and characteristics. (\textit{ii}) Data entanglement. Most evaluations are conducted in real-world scenarios, where the complexity precludes fully controlled experiments to isolate fine-grained factors. (\textit{iii}) Data leakage. Pre-trained LLMs suffer from data leakage and benchmark containment problems, undermining the effectiveness and validity of evaluations.

In this work, we study CoT reasoning by introducing a data distribution lens. Specifically, we hypothesize that CoT reasoning reflects a structured inductive bias learned from in-distribution data, enabling models to conditionally generate reasoning trajectories that approximate those observed during training. As such, the effectiveness of CoT reasoning is fundamentally governed by the nature and degree of distribution discrepancy between training data and test queries. 
Guided by this lens, we revisit existing NLP tasks and identify three primary axes along which distribution shifts may occur: \textit{task} (i.e., unseen task structures), \textit{length} (i.e., different text lengths and reasoning lengths), and \textit{format} (i.e., query format variants).

To tackle the issue of evaluations and validate our hypothesis, we further introduce \methodName{}, an abstract, controllable, and clean environment. \methodName{} provides an abstract representation system that distills various real-world NLP tasks into key components: \emph{atoms} (i.e., token space), \emph{elements} (i.e., text space), and \emph{transformations} (i.e., operation space). By varying these components, we curate data that exhibits various distribution discrepancies, naturally achieving full and fine-grained control over the entire evaluation pipeline. Later, we train models \emph{from scratch} to avoid data leakage and employ controlled experiments to rigorously test our hypotheses.

Our findings reveal that CoT reasoning works effectively when applied to (near) in-distribution data, but becomes fragile and prone to failure even under moderate distribution shifts. In some cases, LLMs generate fluent yet logically inconsistent reasoning steps. The results suggest that what appears to be structured reasoning can be a mirage, emerging from memorized or interpolated patterns in the training data rather than logical inference. Our contributions can be summarized as follows:

\begin{itemize}
    \renewcommand{\labelitemi}{$\star$}
    \item \textbf{Novel perspective.} We propose a \textit{data distribution lens} for CoT reasoning, revealing that its effectiveness arises from structured inductive biases learned from in-distribution data. This lens offers a principled foundation for understanding why and when CoT reasoning succeeds or fails.
    \item \textbf{Controllable environment.} We develop an abstract, fully controllable, and clean environment---\methodName{} that abstracts NLP tasks, enabling systematic analysis of CoT reasoning under distribution discrepancies. \methodName{} can serve as a research platform for probing the intrinsic behavior of LLMs and facilitating the discovery of scientific principles.
    \item \textbf{Rigorous investigation.} Guided by the data distribution lens, we dissect the CoT reasoning via three dimensions: \textit{task}, \textit{length}, and \textit{format}. Later, we curate data that reflects fine-grained factors in each dimension and conduct controlled experiments to isolate and examine each factor.
    \item \textbf{General validity.} We train and fine-tune hundreds of LLMs with varying sizes (from 62K to 14B), architectures (e.g., GPT, LLaMA, and Qwen), and temperatures (from 1e-5 to 10). The results consistently show that the effectiveness of CoT reasoning varies with the degree of distribution discrepancy, substantiating the generality of the proposed data distribution lens.
\end{itemize}

\section{Related Work}
\subsection{LLM Prompting and CoT}
Chain-of-Thought (CoT) prompting improves large language model performance by eliciting intermediate reasoning steps for complex problems \citep{wei2022chain}. Extensions include zero-shot CoT \citep{kojima2022large}, self-consistency via sampling and voting \citep{wang2023selfconsistency}, and Auto-CoT, which automatically generates reasoning exemplars \citep{zhang2023automatic}. Beyond linear reasoning, Tree-of-Thought enables search over multiple reasoning paths \citep{yao2023tree}, while SymbCoT integrates symbolic representations into CoT \citep{xu2024faithful}. More recent work embeds long-form CoT directly into inference, enabling reflection, error correction, and alternative reasoning strategies \citep{jaech2024openai,team2024qwq,guo2025deepseek,team2025kimi,yeo2025demystifying,chen2025towards}. In this work, we investigate whether CoT reflects genuine reasoning or merely pattern interpolation.

\subsection{Discussion on Illusion of LLM Reasoning}
Recent work questions the robustness and faithfulness of these gains~\citep{stechly2024chain}. A prominent line of research shows that CoT reasoning is highly fragile: semantically irrelevant perturbations, such as distractor phrases or altered symbolic representations, can substantially degrade performance~\citep{mirzadehgsm, tang2023large}. Other studies find that models favor surface-level reasoning patterns over logical validity. Moreover, reasoning performance scales poorly with task difficulty, with models over-elaborating on simple problems and failing on harder ones~\citep{shojaee2025illusion}. Concerns about faithfulness further arise from intervention-based analyses showing that final answers often remain unchanged when intermediate steps are corrupted or removed~\citep{lanham2023measuring}, an effect referred to as the illusion of transparency~\citep{chen2025reasoning,bentham2024chainofthought}. The opposing perspectives on CoT reasoning call for a systematic understanding of why and when CoT reasoning succeeds or fails.

\subsection{OOD Generalization of LLMs}
Out-of-distribution (OOD) generalization remains a central challenge in machine learning~\citep{yang2024generalized,yang2023out,budnikov2025generalization,zhang2024can}. Prior work shows that pre-trained models face challenges in adapting to new settings when prompted to learn novel functions~\citep{wang2024can,garg2022can}. Researchers reveal CoT prompting can partially improve OOD generalization ability, especially for tasks that require long reasoning~\citep{yao2025unveiling,shen2025codi}. However, other work claims such gains are not intrinsic. For example, strong arithmetic generalization emerges only when algorithmic biases are encoded in positional representations~\citep{cho2024position}, and finer-grained CoT supervision during training substantially improves OOD performance~\citep{wang2025chain}. Recent studies further indicate that LLM generalizes reliably when common latent structures are shared across distributions~\citep{wang2025theoretical,li2025training}. In light of these insightful findings, we propose rethinking CoT reasoning through a data distribution lens, dissecting CoT reasoning into \textit{task}, \textit{length}, and \textit{format}, and systematically investigating each via controlled experiments. We further provide a comparison with representative work in Appendix~\ref{app:extended_related_work:comparison}.

\section{The Proposed Data Distribution Lens}
We propose the data distribution lens to understand why and when CoT reasoning succeeds or fails. We hypothesize that \textit{CoT reasoning reflects a structured inductive bias learned from in-distribution data, enabling models to conditionally generate reasoning trajectories that approximate those observed during training. As such, the effectiveness of CoT reasoning is fundamentally governed by the nature and degree of distribution discrepancy between training data and test queries (rather than by model architecture or scale).}

To formalize this view, we first introduce notation for the training and test distributions. Let $\mathcal{D}_{\text{train}}$ denote the training distribution over input-output pairs $(x, y)$, where $x$ represents a reasoning problem and $y$ denotes the solution sequence (including intermediate reasoning traces). During training, the model learns a parametric mapping $f_\theta(x) \approx y$ by minimizing the empirical training risk
\begin{equation}
    \hat R_{\text{train}}(f_\theta)
    = \frac{1}{n} \sum_{i=1}^n \ell\bigl(f_\theta(x_i), y_i\bigr)
\end{equation}
where $(x_i, y_i) \sim \mathcal{D}_{\text{train}}$ are i.i.d.\ samples and
$\ell$ is a loss function (e.g., cross-entropy).
The corresponding \emph{expected} (population) training risk is 
\begin{equation}
    R_{\text{train}}(f_\theta)
    = \mathbb{E}_{(x, y) \sim \mathcal{D}_{\text{train}}}
    \bigl[\ell(f_\theta(x), y)\bigr]
\end{equation}
At inference time, given a test query sampled from a potentially different distribution $\mathcal{D}_{\text{test}}$, the model generates a response. The expected test risk is
\begin{equation}
    R_{\text{test}}(f_\theta)
    = \mathbb{E}_{(x, y) \sim \mathcal{D}_{\text{test}}}
    \bigl[\ell(f_\theta(x), y)\bigr]
\end{equation}

\begin{definition}[Distribution Discrepancy]
\label{def:distribution_discrepancy}
Given training distribution $\mathcal{D}_{\text{train}}$ and test distribution
$\mathcal{D}_{\text{test}}$, we define the distribution discrepancy as
\begin{equation}
    \Delta(\mathcal{D}_{\text{train}}, \mathcal{D}_{\text{test}})
    := \operatorname{TV}(\mathcal{D}_{\text{train}}, \mathcal{D}_{\text{test}})
\end{equation}
where $\operatorname{TV}(P,Q)$ is the total variation distance,
\begin{equation}
    \operatorname{TV}(P,Q)
    := \sup_{A} |P(A) - Q(A)|
    = \frac{1}{2} \int |dP - dQ|
\end{equation}
\end{definition}

\begin{theorem}[Generalization Bound]
\label{thm:generalization_bound}
Assume the loss is bounded, i.e., for all $(x,y)$,
$0 \le \ell(f_\theta(x), y) \le B$.
Let $\{(x_i,y_i)\}_{i=1}^n$ be i.i.d.\ samples from $\mathcal{D}_{\text{train}}$ and
let $\hat R_{\text{train}}(f_\theta)$ be the empirical training risk defined above.
Then for any $\delta \in (0,1)$, with probability at least $1-\delta$ over the draw
of the training sample, the expected test risk satisfies
\begin{equation}
    R_{\text{test}}(f_\theta)
    \! \le \!
    \hat R_{\text{train}}(f_\theta)
    \! + \! 2B \,\Delta(\mathcal{D}_{\text{train}}, \mathcal{D}_{\text{test}}) \! + \! B \sqrt{\frac{\log(1/\delta)}{2n}}.
\end{equation}
The proof is provided in Appendix~\ref{app:gen_bound}.
\end{theorem}

Theorem~\ref{thm:generalization_bound} provides a theoretical foundation for the data distribution lens. Guided by it, we identify three critical dimensions along which distribution shifts can occur: task, length, and format.
\begin{equation}
    \Delta(\mathcal{D}_{\text{train}}, \mathcal{D}_{\text{test}}) = \Phi \left(\Delta_\text{task}, \Delta_\text{length},\Delta_\text{format}\right)
\end{equation}
where $\Phi$ is a monotonically increasing composition function that aggregates all discrepancies. $\Delta_\text{task}$, $\Delta_\text{length}$, and $\Delta_\text{format}$ measure the distribution discrepancy introduced by unseen tasks, various lengths, and prompt format variants.

\begin{figure*}
    \centering
    \includegraphics[width=1\linewidth]{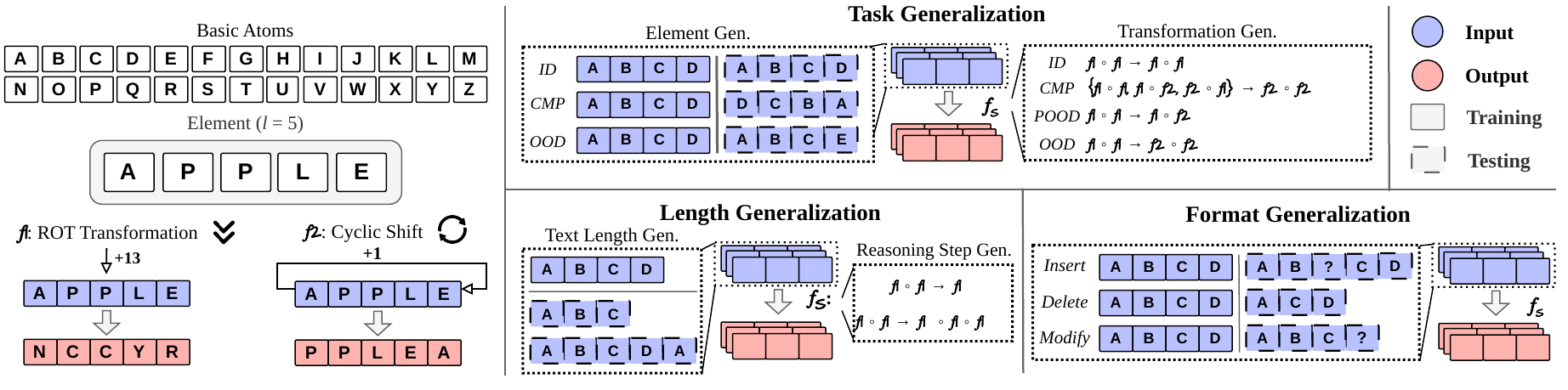}
    \caption{Framework of \methodName{}. \methodName{} provides an abstract representation system that distills various real-world NLP tasks into key components: \emph{atoms}, \emph{elements}, and \emph{transformations}. By varying these components, we curate data that exhibits various distribution discrepancies following \textit{task}, \textit{length}, and \textit{format} generalization. \methodName{} achieves full and fine-grained control over the entire evaluation pipeline. Later, we train models \emph{from scratch} to avoid data leakage and employ controlled experiments to rigorously test the hypotheses.}
    \label{fig:main}
\end{figure*}

\section{DataAlchemy: A Controllable Environment}
\label{sec:dataalchemy}
To empirically validate the data distribution lens, we introduce \methodName{}, an abstract, controllable, and clean environment. It distills real-world NLP tasks into basic \textit{atoms}, \textit{elements}, and \textit{transformations} as illustrated in Figure~\ref{fig:main}.

\subsection{Basic Atoms and Elements}
\label{sec:atom_element_definition}
We abstract tokens in the real-world NLP tasks into basic \textit{atoms} represented by an alphabet of 26 letters $\mathcal{A} = \{\texttt{A}, \texttt{B}, \texttt{C}, \ldots, \texttt{Z}\}$. Based on \textit{atoms}, we further construct an \textit{element} $\mathbf{e}$ as an ordered sequence of atoms with length $l$, reflecting the text space (considering the text consists of tokens):
\begin{equation}
    \mathbf{e} = (a_0, a_1, \ldots, a_{l-1}) \quad \text{where} \ a_i \in \mathcal{A}, \ l \in \mathbb{Z}^+
\end{equation}
Note that we can construct at most $|\mathcal{A}|^l$ distinct elements, which provides a versatile approach for data curation by manipulating element length $l$.

\subsection{Transformations}
\label{sec:transformations}
Similarly, we abstract operations that LLMs perform on text in the real world (e.g., summarize, paraphrase, and reasoning) as \textit{transformations} that operate on elements $F: \mathbf{e} \rightarrow \hat{\mathbf{e}}$. In this work, we mainly instantiate two fundamental \textit{transformations}: the ROT Transformation and the Cyclic Position Shift. Additional \textit{transformations} are considered in the Appendix~\ref{app:quantitative:transformation} to avoid bias. To formally define the \textit{transformations}, we introduce a bijective mapping $\phi: \mathcal{A} \to \mathbb{Z}_{26}$, where $\mathbb{Z}_{26} = \{0, 1, \ldots, 25\}$, such that $\phi(c)$ maps a character to its zero-based alphabetical index.
\begin{definition}[ROT Transformation]
\label{def:rot}
Given an element $\mathbf{e} = (a_0, \ldots, a_{l-1})$ and a rotation parameter $n \in \mathbb{Z}$, the ROT Transformation $f_{\text{rot}}$ produces an element $\hat{\mathbf{e}} = (\hat{a}_0, \ldots, \hat{a}_{l-1})$. Each atom $\hat{a}_i$ is:
\begin{equation}
    \hat{a}_i = \phi^{-1}((\phi(a_i) + n) \pmod{26})
\end{equation}
This operation cyclically shifts each atom $n$ positions forward in alphabetical order. For example, if $\mathbf{e} = (\texttt{A}, \texttt{P}, \texttt{P}, \texttt{L}, \texttt{E})$ and $n=13$, then $f_{\text{rot}}(\mathbf{e}, 13) = (\texttt{N}, \texttt{C}, \texttt{C}, \texttt{Y}, \texttt{R})$.
\end{definition}

\begin{definition}[Cyclic Position Shift]
\label{def:pr}
Given an element $\mathbf{e} = (a_0, \ldots, a_{l-1})$ and a shift parameter $n \in \mathbb{Z}$, the Cyclic Position Shift $f_{\text{pos}}$ produces an element $\hat{\mathbf{e}} = (\hat{a}_0, \ldots, \hat{a}_{l-1})$. Each atom $\hat{a}_i$ is defined by a cyclic shift of indices:
\begin{equation}
    \hat{a}_i = a_{(i + n) \pmod{l}}
\end{equation}
This transformation cyclically shifts the positions of the atoms within the sequence by $n$ positions to the left. For instance, if $\mathbf{e} = (\texttt{A}, \texttt{P}, \texttt{P}, \texttt{L}, \texttt{E})$ and $n=1$, then $f_{\text{pos}}(\mathbf{e}, 1) = (\texttt{P}, \texttt{P}, \texttt{L}, \texttt{E}, \texttt{A})$.
\end{definition}

\begin{definition}[Generalized Compositional Transformation]
\label{def:gct}
To model multi-step reasoning, we define a compositional transformation as the successive application of a sequence of operations. Let $S = (f_1, f_2, \ldots, f_{k})$ be a sequence of operations, where each $f_i$ is one of the fundamental transformations $\mathcal{F}=\{f_{\text{rot}}, f_{\text{pos}}\}$ with its respective parameters. The compositional transformation $f_{\text{S}}$ for the sequence $S$ is the function composition:
\begin{equation}
    f_{\text{S}} = f_{1} \circ f_{2} \circ \cdots \circ f_k
\end{equation}
The resulting element $\hat{\mathbf{e}}$ is obtained by applying the operations sequentially to an initial element $\mathbf{e}$:
\begin{equation}
    \hat{\mathbf{e}} = f_{k}(f_{k-1}(\ldots(f_1(\mathbf{e}))\ldots))
\end{equation}
\end{definition}
This design enables the construction of arbitrary transformations with the type, parameters, order, and length. At the same time, we can naturally acquire the CoT reasoning step by decomposing the intermediate process:

\begin{equation}
\underbrace{f_{\text{S}}(\mathbf{e}):}_{\text{Query}} \quad 
\underbrace{\mathbf{e} \xrightarrow{f_1} \mathbf{e}^{(1)} \xrightarrow{f_2} \mathbf{e}^{(2)} \cdots \xrightarrow{f_{k-1}} \mathbf{e}^{(k-1)} \xrightarrow{f_k}}_{\text{Reasoning traces}} \underbrace{\boxed{\hat{\mathbf{e}}}}_{\text{Answer}}
\end{equation}

Illustrative examples of atoms, elements, and transformations are detailed in Appendix~\ref{app:dataalchemy}.

\subsection{Environment Setting}
Through systematic manipulation of elements and transformations, \methodName{}, we can train and probe various LLMs under various tasks, lengths, and format distributions. In the controlled experiment, we employ decoder-only LLMs with GPT and LLaMA architectures and parameter sizes ranging from 62K to 3B when training from scratch. In the real-world experiments, we utilize two state-of-the-art (SOTA) LLMs:  LLaMA3-8B~\citep{dubey2024LLaMA} and Qwen3-14B-Instruct~\citep{yang2025qwen3}. We construct elements with 2 to 6 basic atoms, which produce 676 to 308,915,776 data samples. We initialize the two transformations $f_1 = f_{\text{rot}}(e,13)$ and $f_2 = f_\text{pos}(e,1)$. We consider both hard metrics, i.e., exact match rate, and soft metrics, i.e., Levenshtein distance (edit distance)~\citep{yujian2007normalized}, and BLEU score~\citep{papineni2002bleu} for evaluation. To enable a fine-grained analysis, we evaluate reasoning traces, the final answer, and the full chain in the LLM response. Detailed environment setting and implementation are provided in Appendix~\ref{app:exp}.

\section{Task Generalization}
\label{sec:task_generalization}
\begin{wrapfigure}{!ht}{0.45\textwidth}
    \centering
    \vspace{-4mm}
    \includegraphics[width=1\linewidth]{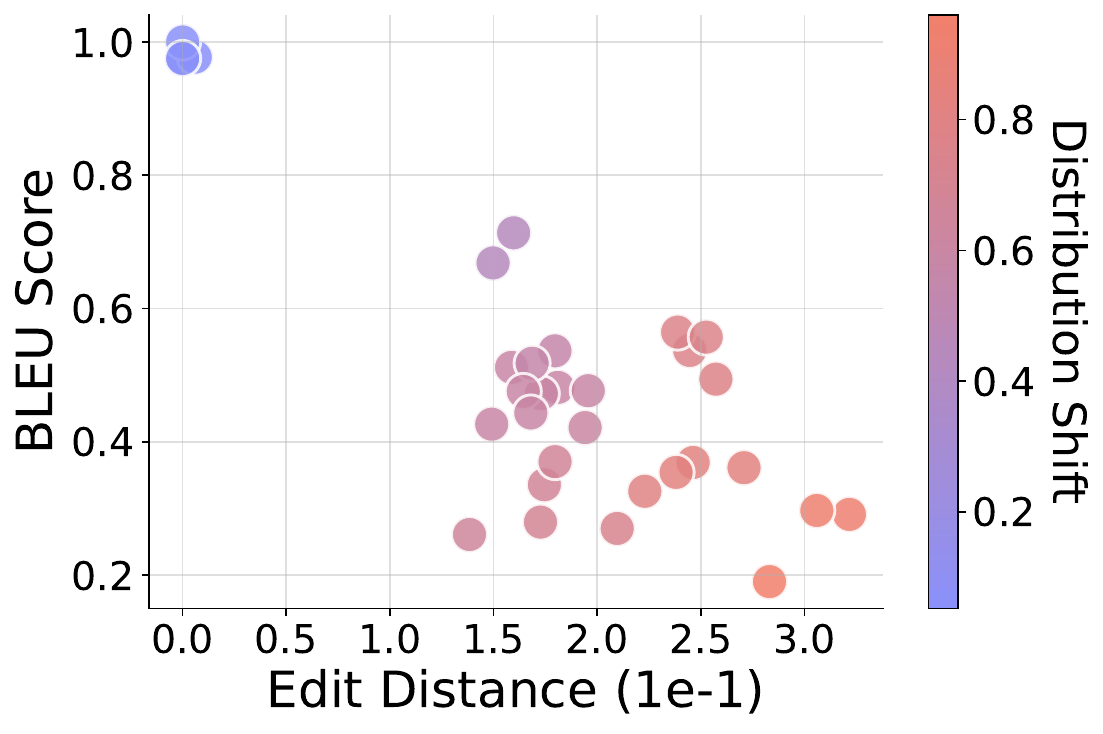}
    \caption{Transformation generalization under different distribution discrepancies. The efficacy of CoT reasoning decreases as task distribution discrepancy increases.}
    \label{fig:transformation_distribution}
\end{wrapfigure}
To investigate the extent to which CoT reasoning can handle \textit{tasks} under various distribution discrepancies, we design task generalization experiments. As we discussed in Section~\ref{sec:dataalchemy}, we decompose tasks into a combination of various \textit{transformations} and \textit{elements}. Therefore, we consider task generalization from two dimensions: transformation generalization and element generalization.

\subsection{Transformation Generalization}
\noindent\textbf{Experiment setup.} To formulate different distribution discrepancies for task generalization, we design the following progressive scenarios based on the proposed measurement (detailed in Appendix~\ref{app:task_distribution}). (\textit{i}) In-Distribution (ID). The transformations in the test set are identical to those observed during training, e.g., $f_1 \circ f_1 \rightarrow f_1 \circ f_1$. (\textit{ii}) Composition (CMP). Test samples comprise novel compositions, where basic transformations are observed during training, e.g., ${f_1 \circ f_1, f_1 \circ f_2, f_2 \circ f_1} \rightarrow f_2 \circ f_2$. (\textit{iii}) Partial Out-of-Distribution (POOD): Test queries include compositions involving both seen and unseen basic transformations, e.g., $f_1 \circ f_1 \rightarrow f_1 \circ f_2$. (\textit{iv}) Out-of-Distribution (OOD). The test set contains entirely novel transformations (compositions) in training, e.g., $f_1 \circ f_1 \rightarrow f_2 \circ f_2$. The illustrative examples for transformation generalization under different scenarios are provided in Appendix~\ref{app:illustration_of_generalization:transformation}.

\begin{table}[!htbp]
\centering
\caption{Full chain evaluation under different scenarios on transformation generalization.}
\label{tab:transformation_full_chain}
\resizebox{0.5\linewidth}{!}{%
\begin{tabular}{@{}l|c|c|c@{}}
\toprule
\textbf{Scenarios} & \textbf{Exact Match (\%)} & \textbf{Edit Distance} & \textbf{BLEU Score} \\
\midrule
ID & 100.00 & 0 & 1 \\
CMP & 0.01 & 0.1326 & 0.6867 \\
POOD & 0.00 & 0.1671 & 0.4538 \\
OOD & 0.00 & 0.2997 & 0.2947 \\
\bottomrule
\end{tabular}%
}
\end{table}

\noindent\textbf{Findings.} Figure~\ref{fig:transformation_distribution} illustrates the performance of the full chain under different distribution discrepancies. 

We can observe that, in general, the effectiveness of CoT reasoning decreases as the distribution discrepancy increases, which directly validates the data distribution lens. As shown in Table~\ref{tab:transformation_full_chain}, CoT reasoning achieves satisfactory performance in the ID (exact match: 100\%) scenario, while it degrades in CMP (0.01\%), POOD (0\%), and OOD (0\%) scenarios. Diving into fine-grained analysis, as demonstrated in Table~\ref{tab:transformation_performance}, we find that the success of CoT reasoning is attributed to the replicating pattern in the training data, as indicated by the inconsistency in reasoning and answers. For instance, when an unseen transformation $f_1 \circ f_1$ is present, LLMs attempt to generalize based on the most similar transformation (i.e., $f_1 \circ f_2$) seen during training, which leads to correct reasoning paths yet incorrect answers. Due to the commutativity of the transforms, generalization from $f_1 \circ f_2$ to $f_2 \circ f_1$ or vice versa allows LLMs to produce incorrect paths yet correct answers, which reflects the unfaithfulness and pattern-matching nature of CoT reasoning. Additional analysis and illustrative examples are provided in Appendix~\ref{app:quantitative:transformation} and~\ref{app:qualitative:task_generalization}.

\begin{table}[!htbp]
\centering
\caption{Fine-grained analysis for CoT reasoning on transformation generalization based on exact match.}
\label{tab:transformation_performance}
\resizebox{0.6\linewidth}{!}{%
\begin{tabular}{@{}l|c|c|c@{}}
\toprule
\textbf{Transformation (Train $\rightarrow$ Test)} & \textbf{Reasoning} & \textbf{Answer} & \textbf{Full Chain} \\
\midrule
$\{f_1 \circ f_1, f_1 \circ f_2, f_2 \circ f_1\} \rightarrow f_2 \circ f_2$ & 100.00 & 0.01 & 0.01 \\
$\{f_1 \circ f_2, f_2 \circ f_1, f_2 \circ f_2\} \rightarrow f_1 \circ f_1$ & 100.00 & 0.01 & 0.01 \\
$f_1 \circ f_2 \rightarrow f_2 \circ f_1$ & 0.00 & 100.00 & 0.00 \\
$f_2 \circ f_1 \rightarrow f_1 \circ f_2$ & 0.00 & 100.00 & 0.00 \\
\bottomrule
\end{tabular}%
}
\end{table}

\noindent\textbf{Experiment setup.} To further probe \textit{when} CoT reasoning can adapt to unseen transformations, we conduct supervised fine-tuning (SFT) experiments to incorporate a portion $\lambda$ of unseen data.

\begin{wrapfigure}{!ht}{0.45\textwidth}
    \centering
    \vspace{-4mm}
    \includegraphics[width=1\linewidth]{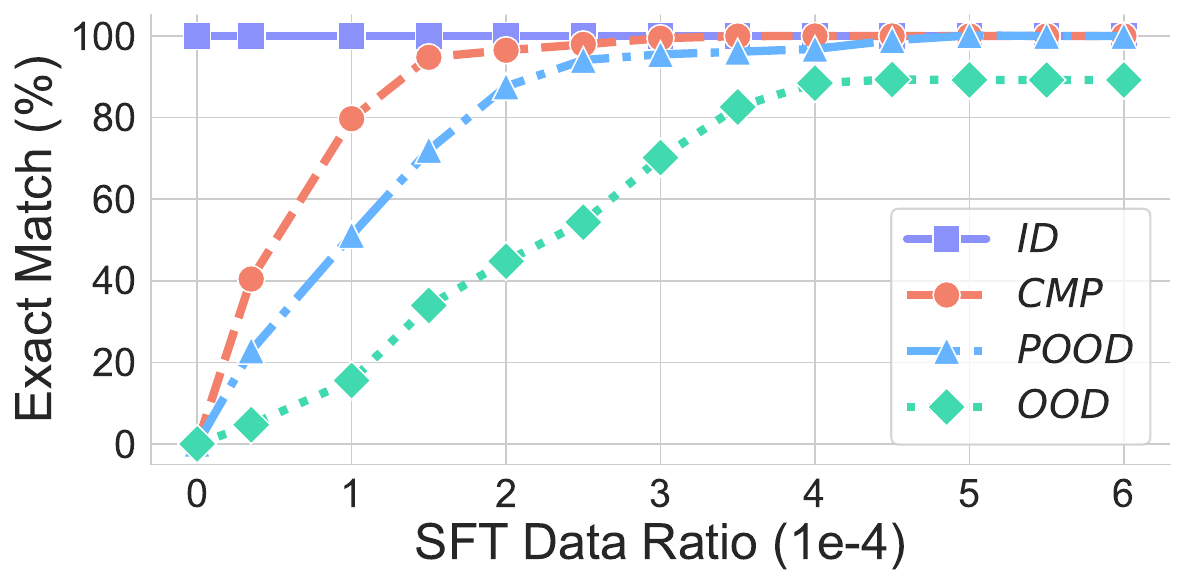}
    \caption{Effectiveness of SFT. A small portion of unseen data helps CoT reasoning to quickly generalize.}
    \vspace{-15mm}
    \label{fig:transformation_sft}
\end{wrapfigure}

\noindent\textbf{Findings.} As shown in Figure~\ref{fig:transformation_sft}, we can find that generally a very small portion ($\lambda=1.5e{-4}$) of data can make the model quickly generalize to unseen transformations. The less discrepancy between the training and testing data, the easier the model can generalize, highlighting the role of similar patterns that appear in the training data.

\subsection{Element Generalization}
Following a pipeline similar to transformation generalization, we investigate how CoT reasoning handles \textit{elements} under various distribution discrepancies. Findings observed also support the proposed data distribution lens. The detailed experiment design and analysis can be found in Appendix~\ref{app:quantitative:element}.

\section{Length Generalization}
\label{sec:length_generalization}
To study how CoT reasoning can operate on varying lengths, we design a length generalization experiment. Following the same intuition as \textit{task generalization}, we also formulate length generalization from two perspectives: \textit{text length} (i.e., element length) \textit{generalization} and \textit{reasoning step} (i.e., transformation length) \textit{generalization}.

\subsection{Text Length Generalization}
\noindent\textbf{Experiment setup.} The text length distribution discrepancy can be measured by element length difference, detailed in Appendix~\ref{app:length_distribution}. We train LLMs on the dataset with text length $l=4$ while fixing other factors and evaluate the performance on a variety of lengths (e.g., from $l=2$ to $l=6$). We provide illustrative examples for text length generalization in Appendix~\ref{app:illustration_of_generalization:text_length}.
\begin{wrapfigure}{!ht}{0.45\textwidth}
    \centering
    \vspace{-4mm}
    \includegraphics[width=1\linewidth]{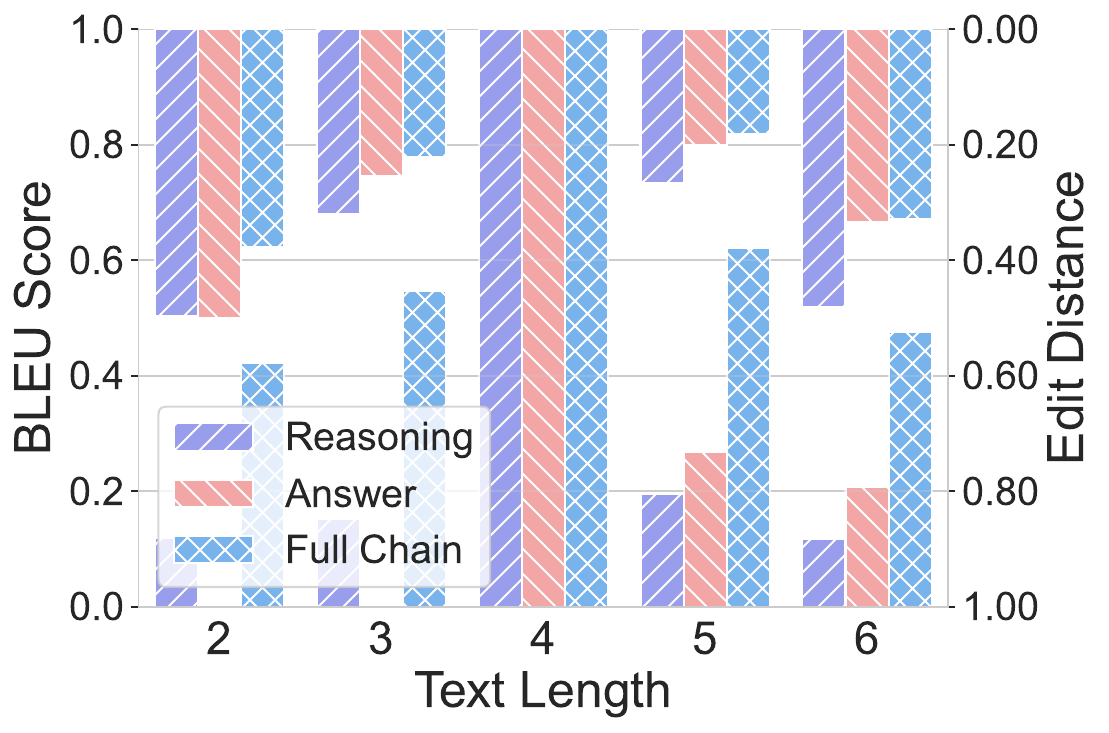}
    \caption{Text length generalization under distribution discrepancies. Increasing distribution shifts in the text length lead to degraded CoT reasoning performance.}
    \label{fig:text_length}
    \vspace{-6mm}
\end{wrapfigure}

\noindent\textbf{Findings.} Figure~\ref{fig:text_length} shows that CoT reasoning produces excellent results under in-distribution scenarios ($l=4$), while its performance degrades as discrepancies in the text length distribution increase, which confirms the data distribution lens. When we further analyze the exact match in Table~\ref{tab:text_length}, CoT reasoning fails to directly generate test cases for those lengths, even with a mild distribution shift (e.g., $l=3$ or $l=5$). Examples in Appendix~\ref{app:text_length_example} indicate that LLMs attempt to produce CoT reasoning with the same length as the training data by adding or removing tokens when processing unseen text length. We further consider the effect of different padding strategies in Appendix~\ref{app:quantitative:text_length}.

\subsection{Reasoning Step Generalization}
\noindent\textbf{Experiment setup.} Reasoning steps are determined by the number of basic \textit{transformations} $k$ in the \textit{compositional transformation}. We mix the data with various reasoning steps (e.g., $k=1,2,3$). By adjusting the mix ratio while maintaining the data size, we create different distribution discrepancies. Examples of reasoning step generalization are detailed in Appendix~\ref{app:illustration_of_generalization:reasoning_step}.

\begin{figure}[!th]
    \centering
    \includegraphics[width=0.49\linewidth]{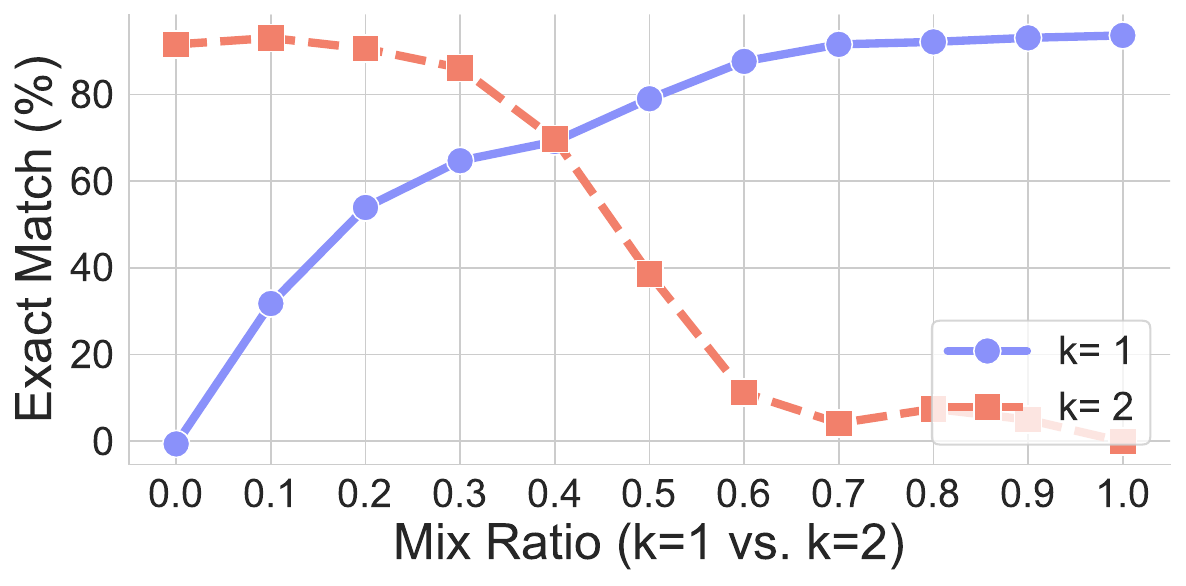}
    \includegraphics[width=0.49\linewidth]{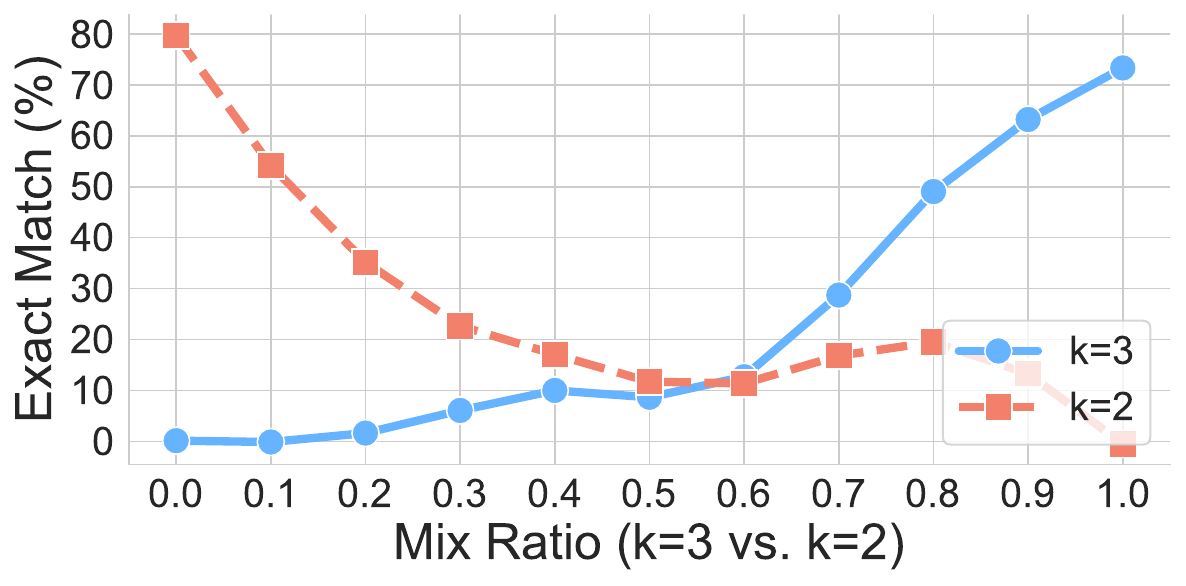}
    \caption{Reasoning step generalization under distribution discrepancies. Performance of CoT reasoning systematically varies with training data components.}
    \label{fig:reasoning_step}
\end{figure}

\noindent\textbf{Findings.} As showcased in Figure~\ref{fig:reasoning_step}, when we adjust the component of training data, the performance of CoT changes accordingly. For instance, increasing the ratio of $k=1$ data will enhance performance on one-step reasoning while compromising two-step reasoning, which supports our hypothesis. Notably, considering extreme cases where the mix ratio is 0 or 1.0, the CoT reasoning achieves good performance in the covered reasoning step but fails to generalize to unseen cases, indicating its fragility when encountering distribution shifts.

\section{Format Generalization}
\label{sec:format_generalization}
To research the robustness of CoT reasoning when surface-level variations appear in test queries, we design the \textit{format generalization}.

\begin{figure}[!th]
    \centering
    \includegraphics[width=0.49\linewidth]{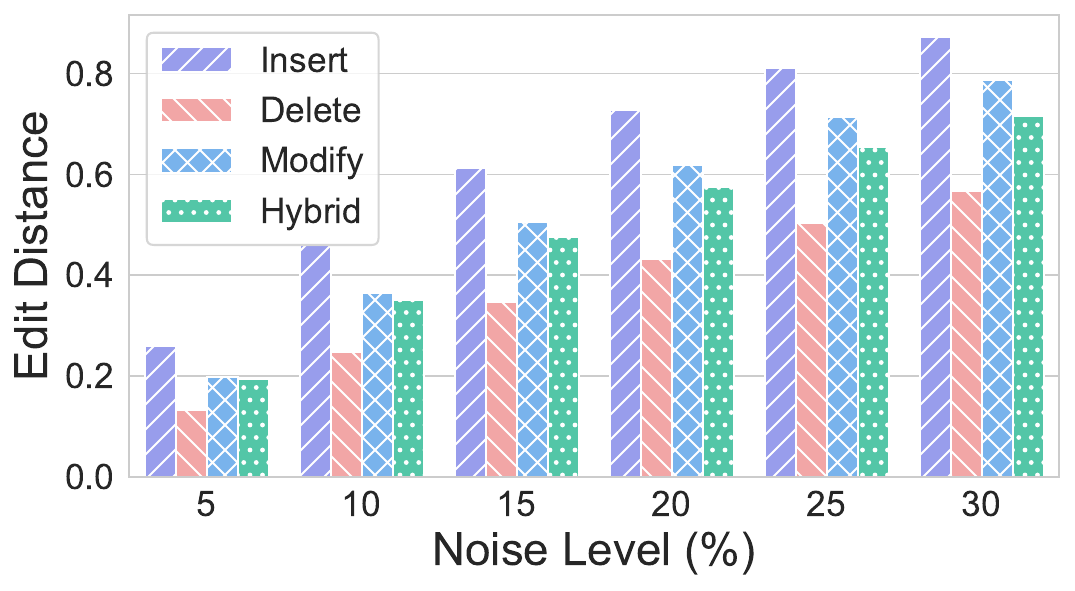}
    \includegraphics[width=0.49\linewidth]{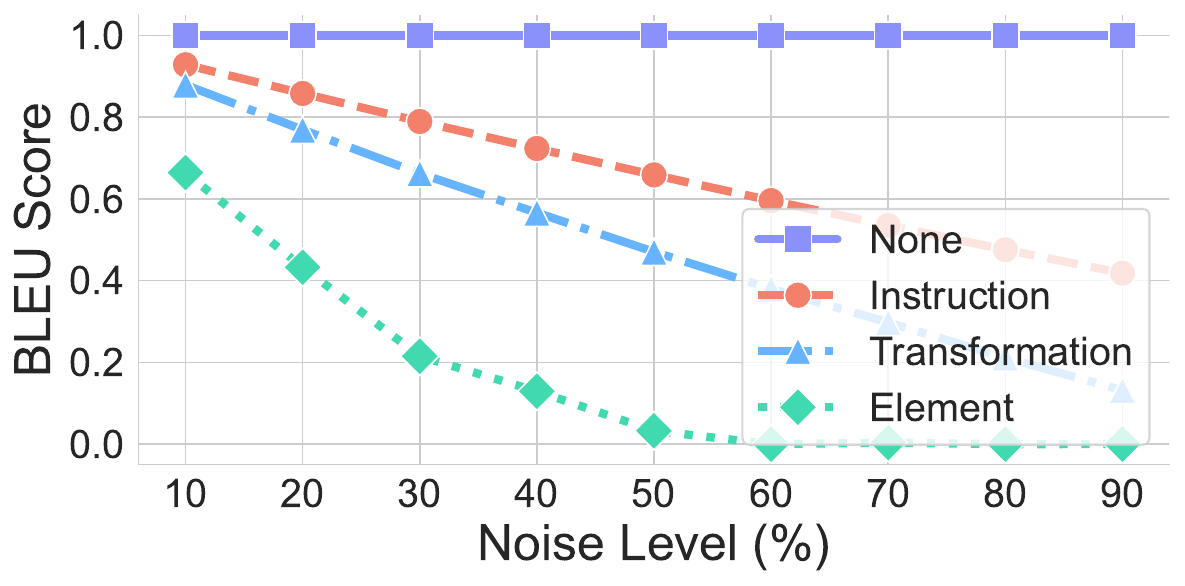}
    \caption{Format generalization under distribution discrepancies. Testing performance degrades with various noise levels and in different applied areas.}
\label{fig:format_generalization}
\end{figure}

\noindent\textbf{Experiment setup.} To introduce the distribution discrepancy at a format level, we propose a distribution measurement (detailed in Appendix~\ref{app:format_distribution}) and consider four distinct perturbation modes to simulate a scenario in the real world. (\textit{i}) Insert. One noise token is inserted. (\textit{ii}) Delete. One original token is deleted. (\textit{iii}) Modify. One original token is replaced with a noise token. (\textit{iv}) Hybrid. It combines the above-mentioned perturbation methods. We apply four perturbations with the noise level of $p$ on different areas (e.g., elements, transformations, and instructions) of test queries. We further elaborate on the format generalization using illustrative examples in Appendix~\ref{app:illustration_of_generalization:format}.

\noindent\textbf{Findings.} As observed in Figure~\ref{fig:format_generalization}, introducing perturbation will compromise the effectiveness of CoT reasoning, and the degree depends on the noise level (i.e., distributional shift), which echoes the data distribution lens. Among different perturbation methods, insertion makes the greatest difference. Considering different areas applied, elements, and transformations play an important role, whereas the changes to other tokens have a lesser effect on the results, which aligns with intuition.

\section{Generality of Data Distribution Lens}
\label{sec:model_size}
To probe the generality of the data distribution lens, we design experiments using LLMs with various architectures, sizes, and temperatures.

\subsection{Internal Validity}
\noindent\textbf{Experiment setup.} For rigor, we conduct the experiments of task, length, and format generalization by training LLMs with GPT and LLaMA architectures and sizes ranging from 62K to 3B.

\begin{figure*}[!th]
    \centering
    \includegraphics[width=0.32\linewidth]{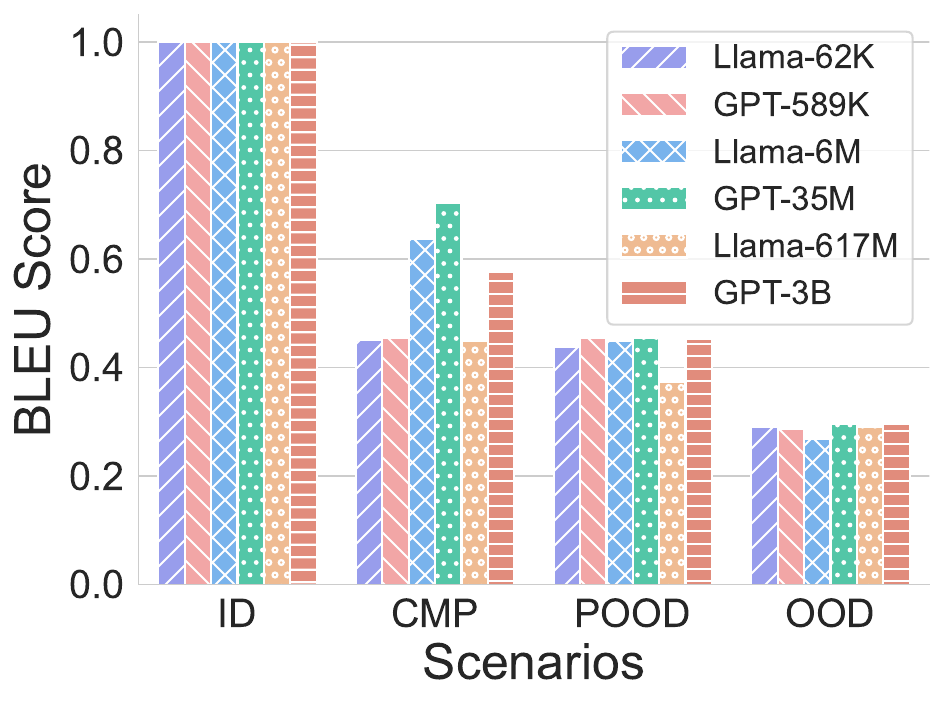}
    \includegraphics[width=0.32\linewidth]{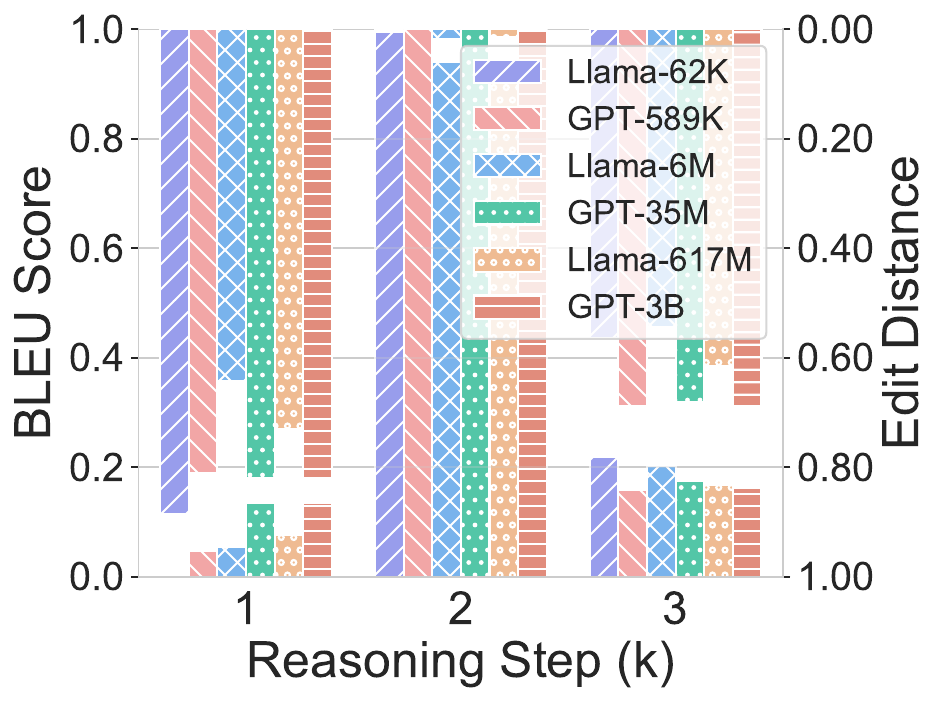}
    \includegraphics[width=0.32\linewidth]{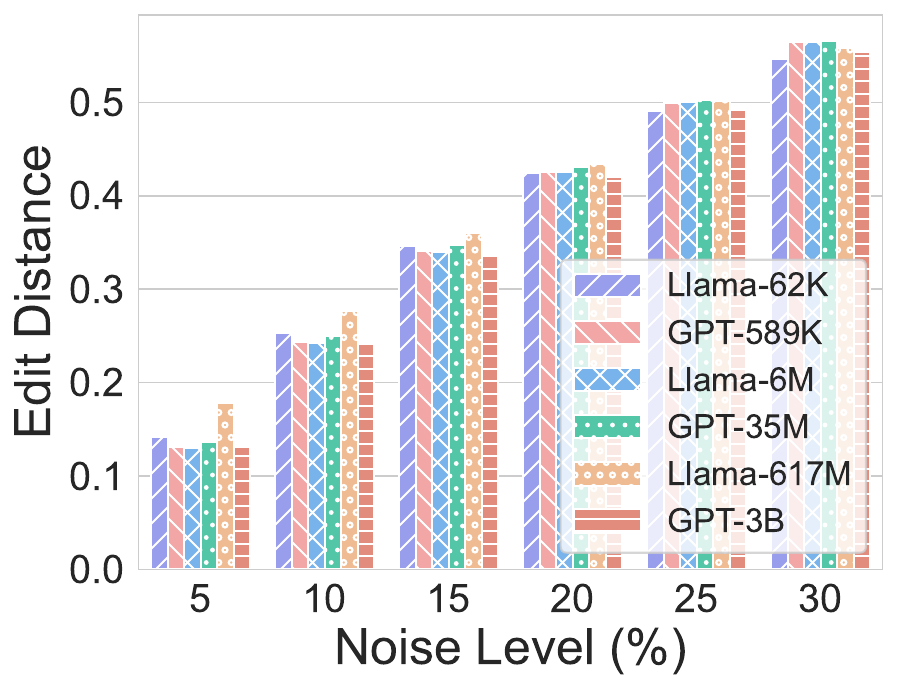}
    \caption{Task, length, and format generalization of LLMs with different settings. The data distribution lens is invariant across LLMs with various sizes and architectures. Results under more settings are provided in Figure~\ref{fig:size_and_architecture_app}.}
    \label{fig:size_and_architecture}
\end{figure*}

\noindent\textbf{Findings.} As illustrated in Figure~\ref{fig:size_and_architecture}, CoT reasoning produced by LLM with different sizes and architectures behaves similarly when encountering distribution shifts on task, length, and format generalization, highlighting the good internal validity. We further study the effect of temperature and the role of SFT with different model sizes in Appendix~\ref{app:quantitative:model_size}.

\subsection{External Validity}
The key to the external validity of the data distribution lens is to identify the distribution discrepancy between training data and test queries, which makes direct evaluation extremely challenging due to the opacity of the training data used by SOTA LLMs. However, this problem can be alleviated if we can curate data unseen during training and then use it to fine-tune LLMs. By interfering with data generated by \methodName{}, where LLMs produce totally random answers, we confirm the validity of the proposed pipelines.

\noindent\textbf{Experiment setup.} We conduct task, length, and format generalization experiments by fine-tuning two SOTA LLMs: LLaMA3-8B~\citep{dubey2024LLaMA} and Qwen3-14B-Instruct~\citep{yang2025qwen3}.

\begin{figure*}[!th]
    \centering
    \includegraphics[width=0.32\linewidth]{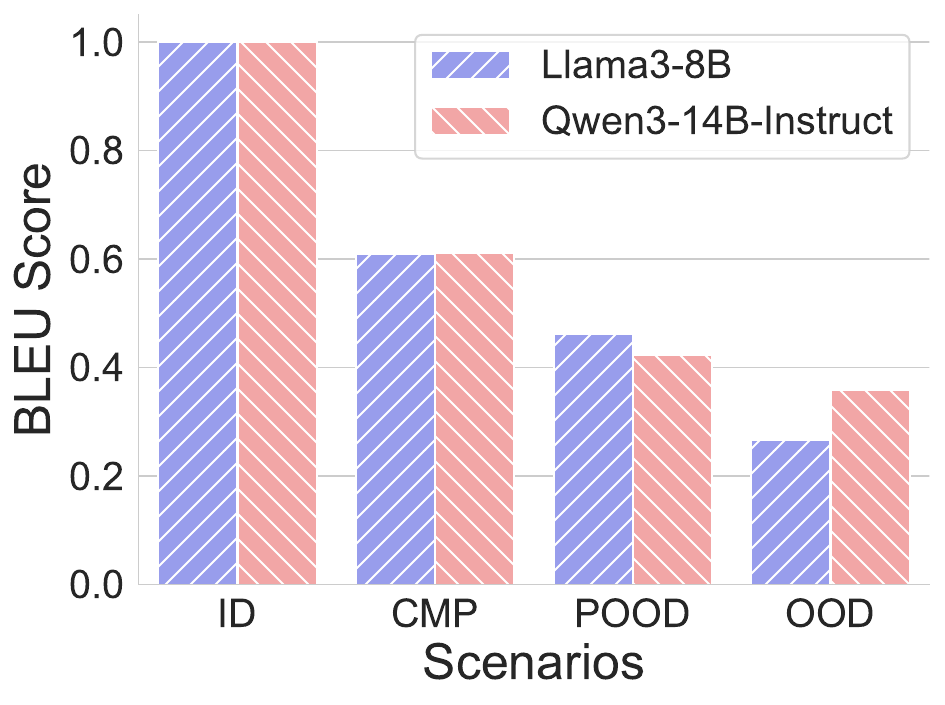}
    \includegraphics[width=0.32\linewidth]{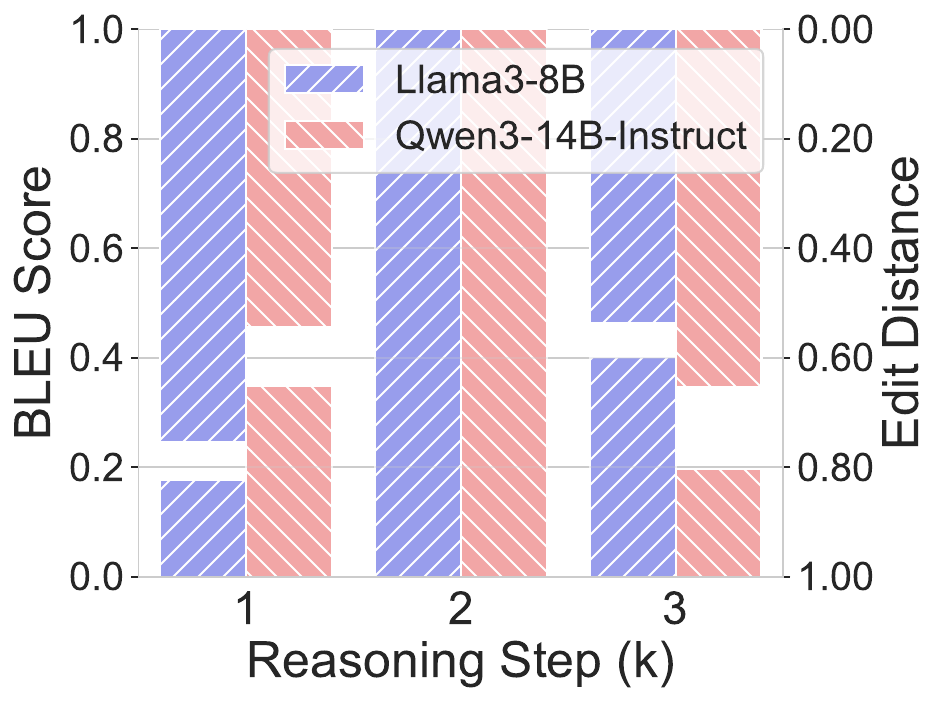}
    \includegraphics[width=0.32\linewidth]{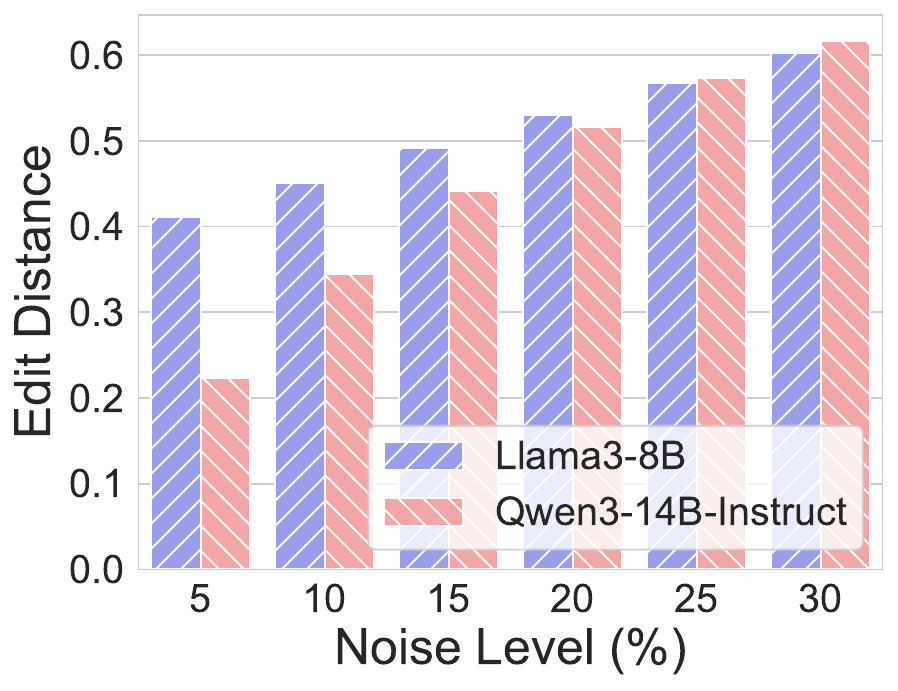}
    \caption{Task, length, and format generalization of SOTA LLMs. The data-distribution lens is valid.}
    \label{fig:real_world}
\end{figure*}

\noindent\textbf{Findings.}
As shown in Figure~\ref{fig:real_world}, the performance of SOTA LLMs exhibits similar trends to pre-trained models in \methodName{} across task, length, and format generalization, indicating the external validity of the distribution lens. Additional results and analysis are provided in Appendix~\ref{app:quantitative:external}.

\section{Discussion and Implication}
Through rigorous experiments, we demonstrate that CoT reasoning is effective when applied to (near) in-distribution data, but becomes fragile and prone to failure even under moderate distribution shifts. What appears to be structured reasoning can be a mirage, emerging from memorized or interpolated patterns in the training data rather than logical inference. Our work carries important implications for both LLM researchers and practitioners, which are further discussed in Appendix~\ref{app:discussion}.

\section{Conclusion}
We examine the CoT reasoning of LLMs through the data distribution lens, revealing that the perceived structured reasoning capability largely arises from inductive biases shaped by in-distribution training data, whose effectiveness is bounded by distribution discrepancies. We propose a fully controllable framework, \methodName{}, and systematically probe CoT reasoning with distribution discrepancies introduced by \textit{task}, \textit{length}, and \textit{format}. Comprehensive experiments confirm that the data distribution is invariant across LLMs with different architectures and sizes. We hope \methodName{} can serve as a platform where researchers can rigorously explore the nature of LLMs, inspiring the discovery of universal principles.

\section*{Limitations}
While our work offers a rigorous, controlled investigation into the nature of Chain-of-Thought (CoT) reasoning, we acknowledge several limitations that provide avenues for future research:

(\textit{i}) Synthetic environment vs. natural language complexity. Our controlled experiments are conducted on the abstract environment \methodName{}, which distills real-world language tasks into symbolic atoms, elements, and transformations. While this abstraction enables full and fine-grained control over distribution factors and avoids data leakage, it may inevitably not fully capture the semantic richness, ambiguity, and compositional diversity present in natural language. While external validity of the proposed data distribution is confirmed by real-world SOTA LLMs, the observed brittleness of CoT reasoning under distribution shifts may manifest more stealthily, sophistically, and task-dependently in more complex real-world settings.

(\textit{ii}) Distribution discrepancy measurement and data opacity. Although we evaluate a wide range of model architectures, sizes, and temperatures, including both models trained from scratch and state-of-the-art pretrained LLMs, the training data distributions of commercial or large proprietary models remain uncovered due to the opaque nature of training data and model weights. As a result, fully estimating the distribution discrepancy between pretraining data and test queries is inherently challenging, limiting the precision with which our data distribution lens can be quantitatively validated in fully realistic and transparent scenarios.

(\textit{iii}) Scope of generalization dimensions. We focused our analysis on three specific dimensions of generalization: task, length, and format. While these cover a broad spectrum of OOD scenarios, we did not explicitly model other forms of distribution shift, such as cross-lingual transfer, multi-modal reasoning, or shifts in cultural context.

\section*{Ethical Considerations}
This work studies the reasoning behavior of large language models and does not involve human subjects, personal data, or user-generated content. All experiments are conducted on synthetic or publicly available benchmarks and models, and models trained from scratch use data generated entirely within the proposed framework, avoiding issues of privacy, consent, or data misuse.

Our findings highlight that CoT reasoning can produce fluent yet logically inconsistent or unfaithful reasoning traces when models are evaluated outside their training distributions. This has ethical implications for the deployment of LLMs in high-stakes applications such as education, healthcare, law, and scientific decision-making, where users may over-trust seemingly coherent reasoning explanations. We emphasize that the presence of a detailed reasoning trace should not be equated with correctness, reliability, or genuine understanding.

By exposing the limitations and fragility of CoT reasoning, this work aims to promote more responsible use of LLMs and encourage the research community to develop evaluation protocols and modeling approaches that better reflect true generalization and reasoning capabilities. We believe that transparency about these limitations is essential for preventing misuse and misinterpretation of LLM-generated reasoning.

\bibliographystyle{abbrvnat}
\nobibliography*
\bibliography{custom}
\appendix
\clearpage
\setcounter{tocdepth}{0}
\addtocontents{toc}{\protect\setcounter{tocdepth}{2}}
\renewcommand{\contentsname}{Appendix Contents}

\tableofcontents
\clearpage

\section{Extended Related Work and Comparison}
\label{app:extended_related_work}
\subsection{LLM Prompting and CoT}
Chain-of-Thought (CoT) prompting revolutionized how we elicit reasoning from large language models by decomposing complex problems into intermediate steps~\citep{wei2022chain}. By augmenting few-shot exemplars with reasoning chains, CoT showed substantial performance gains on various tasks~\citep{xu2024faithful,imani2023mathprompter,wei2022chain}. Building on this, several variants emerged. Zero-shot CoT triggers reasoning without exemplars using instructional prompts~\citep{kojima2022large}, and self-consistency enhances performance via majority voting over sampled chains~\citep{wang2023selfconsistency}. To reduce manual effort, Auto-CoT generates CoT exemplars using the models themselves~\citep{zhang2023automatic}. Beyond linear chains, Tree-of-Thought (ToT) frames CoT as a tree search over partial reasoning paths~\citep{yao2023tree}, enabling lookahead and backtracking. SymbCoT combines symbolic reasoning with CoT by converting problems into formal representations~\citep{xu2024faithful}. Recent work increasingly integrates CoT into the LLM inference process, generating long-form CoTs~\citep{jaech2024openai,team2024qwq,guo2025deepseek,team2025kimi}. This enables flexible strategies like mistake correction, step decomposition, reflection, and alternative reasoning paths~\citep{yeo2025demystifying,chen2025towards}. The success of prompting techniques and long-form CoTs has led many to view them as evidence of emergent, human-like reasoning in LLMs. In this work, we investigate whether CoT reflects genuine reasoning or merely pattern interpolation.

\subsection{Discussion on Illusion of LLM Reasoning}
While Chain-of-Thought prompting has led to impressive gains on complex reasoning tasks, a growing body of work has started questioning the nature of these gains~\citep{stechly2024chain}. One major line of research highlights the fragility of CoT reasoning. Minor and semantically irrelevant perturbations such as distractor phrases or altered symbolic forms can cause significant performance drops in state-of-the-art models~\citep{mirzadehgsm, tang2023large}. Models often incorporate such irrelevant details into their reasoning, revealing a lack of sensitivity to salient information. Other studies show that models prioritize the surface form of reasoning over logical soundness; in some cases, longer but flawed reasoning paths yield better final answers than shorter, correct ones~\citep{bentham2024chainofthought}. Similarly, performance does not scale with problem complexity as expected---models may overthink easy problems and give up on harder ones~\citep{shojaee2025illusion}. Another critical concern is the faithfulness of the reasoning process. Intervention-based studies reveal that final answers often remain unchanged even when intermediate steps are falsified or omitted~\citep{lanham2023measuring}, a phenomenon dubbed the illusion of transparency~\citep{bentham2024chainofthought, chen2025reasoning}. Together, these findings suggest that LLMs are not principled reasoners but rather sophisticated simulators of reasoning-like text. However, a systematic understanding of why and when CoT reasoning succeeds or fails is still a mystery.

\subsection{OOD Generalization of LLMs}
Out-of-distribution (OOD) generalization, where test inputs differ from training data, remains a key challenge in machine learning, particularly for large language models (LLMs)~\citep{yang2024generalized,yang2023out,budnikov2025generalization,zhang2024can}. Recent studies show that LLMs prompted to learn novel functions often revert to similar functions encountered during pretraining~\citep{wang2024can,garg2022can}. Likewise, LLM generalization frequently depends on mapping new problems onto familiar compositional structures~\citep{song2025out}. CoT prompting improves OOD generalization~\citep{wei2022chain}, with early work demonstrating length generalization for multi-step problems beyond training distributions~\citep{yao2025unveiling,shen2025codi}. However, this ability is not inherent to CoT and heavily depends on model architecture and training setups. For instance, strong generalization in arithmetic tasks was achieved only when algorithmic structures were encoded into positional encodings~\citep{cho2024position}. Similarly, finer-grained CoT demonstrations during training boost OOD performance, highlighting the importance of data granularity~\citep{wang2025chain}. Theoretical and empirical evidence show that CoT generalizes well only when test inputs share latent structures with training data; otherwise, performance declines sharply~\citep{wang2025theoretical,li2025training}. Despite its promise, CoT still struggles with genuinely novel tasks or formats. In light of these insightful findings, we propose rethinking CoT reasoning through a data distribution lens: decomposing CoT into \textit{task}, \textit{length}, and \textit{format} generalization, and systematically investigating each via controlled experiments.

\begin{figure*}[!th]
    \centering
    \includegraphics[width=0.7\linewidth]{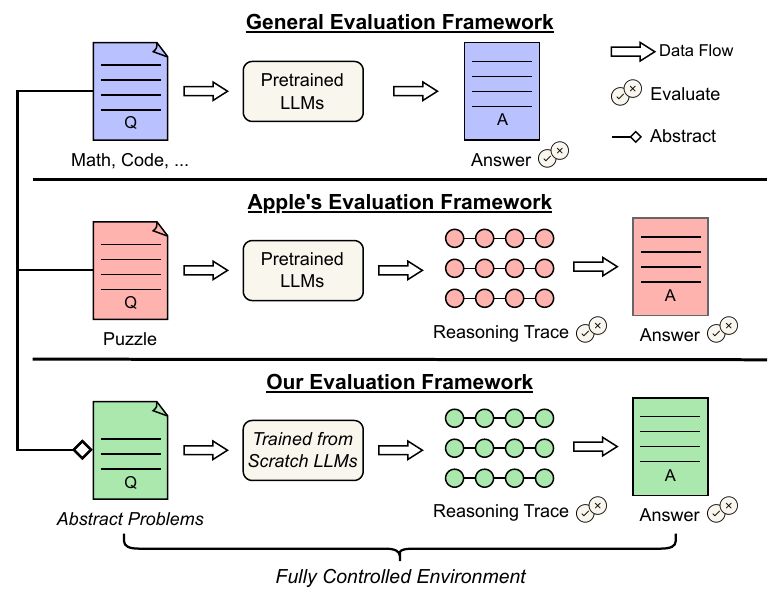}
    \caption{Comparison with representative evaluation method. \methodName{} distills real-world NLP problems, allowing training LLMs from scratch to avoid data leakage issues and study CoT reasoning through rigorous controlled experiments.}
    \label{fig:comparison}
\end{figure*}

\subsection{Comparison with Representative Work}
\label{app:extended_related_work:comparison}
Recent research attempts to create a fine-grained evaluation by examining both the CoT reasoning trace and the final answer through controlled experiments. However, they suffer from: (\textit{i}) Narrowly defined settings: focusing on specific tasks, domains, or LLMs, thereby overlooking the common characteristics across tasks and LLMs. (\textit{ii}) Data entanglement: most evaluations are conducted on real-world tasks and models, where the complexity precludes fully controlled experiments. (\textit{iii}) Data leakage: LLM training makes use of all available data, including benchmarks, undermining the effectiveness and validity of evaluations, which is illustrated in Figure~\ref{fig:comparison}.

To scientifically and rigorously examine CoT reasoning, a new evaluation framework is required. An ideal framework should satisfy the following criteria: (\textit{i}) Abstract representation: it should abstract and unify diverse NLP tasks and LLMs while retaining their essential properties. (\textit{ii}) Fully controlled experiment: it should enable a full and fine-grained control of both tasks (e.g., complexity) and LLMs (e.g., size and architecture), enabling rigorous study of different factors through controlled experiments. (\textit{iii}) Training from scratch: it should offer scalable structural data to train LLMs from scratch, mitigating the data leakage and providing clean evaluations.

Additionally, our proposed framework enables the study of why and when CoT reasoning succeeds or fails.

\section{Illustrative Examples}
\label{app:dataalchemy}
To complement Section~\ref{sec:dataalchemy} and Sections~\ref{sec:task_generalization}--\ref{sec:format_generalization}, this appendix walks through concrete instantiations of \methodName{}'s components and of the training/test templates used for each generalization dimension. Running all examples on the seed element \texttt{APPL} makes the effect of each operator, and the exact distribution shift induced by each template, directly comparable.

\subsection{Components: Atoms, Elements, and Transformations}
\subsubsection{Elements}
As defined in Section~\ref{sec:atom_element_definition}, an element $\mathbf{e}$ is defined as an ordered sequence of atoms. The following is an element with 3 atoms:
\tcbset{
    promptstyle/.style={
        colback=gray!5,     
        colframe=black,      
        fonttitle=\bfseries, 
        boxrule=0.2mm,       
        sharp corners,       
        enhanced,
        breakable,
        width=1\linewidth
    }
}
\begin{tcolorbox}[promptstyle]
\ttfamily
A P P
\end{tcolorbox}
The length of the elements can vary, and the following is an element with 4 atoms:
\begin{tcolorbox}[promptstyle]
\ttfamily
A P P L
\end{tcolorbox}
\subsubsection{Transformations}
A transformation is defined as an operator acting on the elements, as detailed in Section~\ref{sec:transformations}. In the appendix, we additionally consider a third transformation $f_3$ alongside the two fundamental transformations $f_1$ and $f_2$ introduced in the main paper. For brevity in subsequent illustrations and tables, we abbreviate the fundamental transformations as $f_1 := f_{\text{rot}}$ and $f_2 := f_{\text{pos}}$, and additionally introduce $f_3$ as a sequence-reversal operator (defined below).

Specifically, $f_1$ represents an element-wise ROT-13 operation applied to the alphabet. The following example demonstrates the application of $f_1$ to the input \texttt{APPL}:
\begin{tcolorbox}[promptstyle]
\ttfamily
A P P L [F1] <answer> N C C Y
\end{tcolorbox}
$f_2$ denotes a cyclic positional shift by one step. The following example illustrates the result of applying $f_2$ to the sequence \texttt{APPL}:
\begin{tcolorbox}[promptstyle]
\ttfamily
A P P L [F2] <answer> P P L A
\end{tcolorbox}
$f_3$ denotes the sequence reversal operator. Formally, given $\mathbf{e} = (a_0, \ldots, a_{l-1})$, the reversal operator produces $\hat{\mathbf{e}} = (\hat{a}_0, \ldots, \hat{a}_{l-1})$ with $\hat{a}_i = a_{l-1-i}$. The following example displays the result of applying $f_3$ to \texttt{APPL}:
\begin{tcolorbox}[promptstyle]
\ttfamily
A P P L [F3] <answer> L P P A
\end{tcolorbox}
\subsubsection{Compositional Transformations}
Compositional transformations refer to imposing multiple transformations sequentially on an element.
Below are the examples of transformations $f_1$ and $f_2$:
\begin{tcolorbox}[promptstyle]
\ttfamily
A P P L [F1] [F2] 
<think> N C C Y [F2] 
<answer> C C Y N
\end{tcolorbox}

\subsection{Task Generalization Templates}
\label{app:illustration_of_generalization}
\label{app:illustration_of_generalization:task}
As shown in Fig.~\ref{fig:main}, a task is defined through an element and a transformation, and hence task generalization decomposes into generalization on elements and on transformations, respectively.
\subsubsection{Transformation Generalization}
\label{app:illustration_of_generalization:transformation}
We consider the transformation generalization on four aspects: in distribution (ID), compositional (CMP), partially out of distribution (POOD), and out of distribution (OOD).
We use the element \texttt{APPL} to further demonstrate the four aspects in detail:
\paragraph{In distribution} refers to the scenario that the test transformations are identical to the training ones:
\begin{tcolorbox}[promptstyle]
\ttfamily
Training: \\
A P P L [F1] [F2] 
<think> N C C Y [F2] 
<answer> C C Y N \\
Test: \\
A P P L [F1] [F2] <think>
\end{tcolorbox}
\paragraph{Compositional} refers to the scenario where the test transformations are the composition of the training ones.
\begin{tcolorbox}[promptstyle]
\ttfamily
Training: \\
A P P L [F1] [F2] 
<think> N C C Y [F2] 
<answer> C C Y N \\
A P P L [F2] [F1] <think> P P L A [F1] <answer> C C Y N \\
A P P L [F2] [F2] <think> P P L A [F2] <answer> P L A P \\
Test: \\
A P P L [F1] [F1] <think>
\end{tcolorbox}
\paragraph{Partially out of distribution} refers to the scenario where part of the compositional test transformations are seen during training, while the entire compositional transformation is different from training.
\begin{tcolorbox}[promptstyle]
\ttfamily
Training: \\
A P P L [F1] [F1]
<think> N C C Y [F1]
<answer> A P P L \\
Test: \\
A P P L [F1] [F2] <think>
\end{tcolorbox}
\paragraph{Out of distribution} refers to the scenario where none of the compositional test transformations are seen during training.
\begin{tcolorbox}[promptstyle]
\ttfamily
Training: \\
A P P L [F2] [F2] <think> P P L A [F2] <answer> P L A P \\
Test: \\
A P P L [F1] [F1] <think>
\end{tcolorbox}
\subsubsection{Element Generalization}
\label{app:illustration_of_generalization:element}
We consider the element generalization on three aspects: in distribution (ID), compositional (CMP), and out of distribution (OOD).
We use the element \texttt{APPL} to further demonstrate the three aspects in detail:
\paragraph{In distribution} refers to the scenario that the test elements are identical to the training ones:
\begin{tcolorbox}[promptstyle]
\ttfamily
Training: \\
A P P L [F1] [F2] 
<think> N C C Y [F2] 
<answer> C C Y N \\
Test: \\
A P P L [F1] [F2] <think>
\end{tcolorbox}
\paragraph{Compositional} refers to the scenario where the test elements have the same atoms as the training ones but with a different order.
\begin{tcolorbox}[promptstyle]
\ttfamily
Training: \\
A P P L [F1] [F2] 
<think> N C C Y [F2] 
<answer> C C Y N \\
Test: \\
P A L P [F1] [F2] <think>
\end{tcolorbox}
\paragraph{Out of distribution} refers to the scenario where novel atoms appear in the test elements.
\begin{tcolorbox}[promptstyle]
\ttfamily
Training: \\
A P P L [F1] [F2] 
<think> N C C Y [F2] 
<answer> C C Y N \\
Test: \\
A P P Y [F1] [F2] <think>
\end{tcolorbox}
\subsection{Length Generalization Templates}
\label{app:illustration_of_generalization:length}
Similarly, length generalization decomposes into text-length generalization and reasoning-step generalization.
\subsubsection{Text Length Generalization}
\label{app:illustration_of_generalization:text_length}
Text length generalization involves different lengths of elements in the test set from those in the training set. Still, we use \texttt{APPL} as an example to illustrate text length generalization.
\begin{tcolorbox}[promptstyle]
\ttfamily
Training: \\
A P P L [F2] [F2] <think> P P L A [F2] <answer> P L A P \\
Test: \\
A P P L E [F1] [F2] <think>\\
Test: \\
A P P [F1] [F2] <think>
\end{tcolorbox}
\subsubsection{Reasoning Step Generalization}
\label{app:illustration_of_generalization:reasoning_step}
Reasoning step generalization refers to the scenario where the number of transformations in the test compositional transformation is different from that in the training compositional transformation. 
\begin{tcolorbox}[promptstyle]
\ttfamily
Training: \\
A P P L [F2] [F2] <think> P P L A [F2] <answer> P L A P \\
Test: \\
A P P L [F2] <think>\\
Test: \\
A P P L [F2] [F2] [F2] <think>
\end{tcolorbox}
\subsection{Format Generalization Templates}
\label{app:illustration_of_generalization:format}
Format generalization refers to scenarios where the input formats are changed during testing. In this work, we consider three different mechanisms by which the format is changed: addition, deletion, and modification.
\textbf{Insert.} It refers to an additional unknown token that is added to the task. The following example shows how the \texttt{APPL} gets changed under addition:
\begin{tcolorbox}[promptstyle]
\ttfamily
Training: \\
A P P L [F2] [F2] <think> P P L A [F2] <answer> P L A P \\
Test: \\
A P <noise> P L [F2] [F2] <think>
\end{tcolorbox}

\textbf{Delete.} It refers to tokens that are deleted during the test. The following example shows how the \texttt{APPL} gets changed under deletion:
\begin{tcolorbox}[promptstyle]
\ttfamily
Training: \\
A P P L [F2] [F2] <think> P P L A [F2] <answer> P L A P \\
Test: \\
A P L [F2] [F2] <think>
\end{tcolorbox}

\textbf{Modify.} It refers to tokens that are replaced by an unknown token during the test. The following example shows how the \texttt{APPL} gets changed under modification:
\begin{tcolorbox}[promptstyle]
\ttfamily
Training: \\
A P P L [F2] [F2] <think> P P L A [F2] <answer> P L A P \\
Test: \\
A <noise> P L [F2] [F2] <think>
\end{tcolorbox}

\section{Theory and Proofs}
To make the data-distribution lens quantitative along each axis, we introduce a discrepancy measure for \emph{task}, \emph{length}, and \emph{format} generalization, each feeding into the distribution-discrepancy term of Theorem~\ref{thm:generalization_bound}. For each axis, we state the corresponding proposition or definition and, where applicable, the accompanying proof or heuristic derivation directly below it, so that each subsection is self-contained. We close with the proof of the main generalization bound.

\subsection{Task Discrepancy Measure}
\label{app:task_distribution}
We decompose tasks into combinations of \textit{transformations} and \textit{elements} (Section~\ref{sec:dataalchemy}), and consider task generalization along two dimensions: transformation generalization and element generalization.

\textbf{Task Generalization Complexity.} Guided by the data distribution lens, we first introduce a measure for generalization difficulty:

\begin{proposition}[Task Generalization Complexity]
\label{prop:tgc}
Let a task configuration $C$ consist of an element $\mathbf{e} = (a_0, \ldots, a_{l-1})$ of length $l$ and a compositional transformation $f_{\text{S}} = (f_1, \ldots, f_k)$ of depth $k$. We define the task complexity score
\begin{equation}
\label{eq:tgc}
\mathcal{T}(C) \;=\; \alpha \sum_{i=0}^{l-1} \mathbb{1}\!\left[a_i \notin \mathcal{E}^i_{\text{train}}\right] + \beta \sum_{j=1}^{k} \mathbb{1}\!\left[f_j \notin \mathcal{F}_{\text{train}}\right] + \gamma\, \mathbb{1}\!\left[f_{\text{S}} \notin \mathcal{P}_{\text{train}}\right],
\end{equation}
where $\alpha, \beta, \gamma > 0$ weight the three novelty axes: position-wise atoms, individual transformations, and full transformation compositions. The sets $\mathcal{E}^i_{\text{train}}$, $\mathcal{F}_{\text{train}} \subseteq \mathcal{F}$, and $\mathcal{P}_{\text{train}}$ denote, respectively, the set of atoms observed at position $i$ in training elements, the set of fundamental transformations seen in training (a subset of the universe $\mathcal{F}$ from Section~\ref{sec:transformations}), and the set of transformation compositions seen in training. The task-axis discrepancy is obtained from $\mathcal{T}(C)$ through a monotone mapping,

\begin{equation}
\label{eq:delta_task}
\Delta_{\text{task}} \;=\; \psi_{\text{task}}\bigl(\mathcal{T}(C)\bigr),
\end{equation}
where $\psi_{\text{task}} : [0,\infty) \to [0,1]$ is monotonically non-decreasing and satisfies $\psi_{\text{task}}(0) = 0$ (zero novelty $\Rightarrow$ zero task-axis discrepancy). The identity choice $\psi_{\text{task}}(x) = x$ (after suitable normalization) is a natural instance; the argument below depends only on monotonicity. Through this mapping, $\Delta_{\text{task}}$ plugs directly into the distribution-discrepancy term of Theorem~\ref{thm:generalization_bound}.
\end{proposition}

We next establish a critical threshold beyond which the probability of correct CoT reasoning decays exponentially in $\mathcal{T}(C)$:

\begin{theorem}[Task Generalization Failure Threshold]
\label{thm:task_failure}
Let $\mathbf{y}^\star(C)$ denote the ground-truth output chain for configuration $C$, and define
\begin{equation}
\pi(C) \;:=\; \Pr \bigl[\,\mathrm{CoT}_\theta(C) = \mathbf{y}^\star(C)\,\bigr],
\end{equation}
the success probability of CoT reasoning on $C$. Under the multiplicative failure model stated below, there exist positive constants $\kappa$ and $\tau$ such that whenever $\mathcal{T}(C) > \tau$,
\begin{equation}
\pi(C) \;\le\; \exp \bigl(-\kappa\,(\mathcal{T}(C) - \tau)\bigr).
\end{equation}
In words, above the threshold $\tau$, $\pi(C)$ is upper-bounded by a function that decays exponentially in $\mathcal{T}(C)$.
\end{theorem}

\label{app:proof_task_failure}
\begin{proof}[Proof of Theorem~\ref{thm:task_failure}]
We prove the bound by multiplicatively composing per-axis degradation factors and then relating their logarithms to $\mathcal{T}$.

Let $C$ denote a task configuration with element $\mathbf{e} = (a_0, \ldots, a_{l-1})$ and compositional transformation $f_{\text{S}} = (f_1, \ldots, f_k)$, and define the novelty events
\begin{equation*}
\mathsf{N}^{\mathsf{a}}_i := \{a_i \notin \mathcal{E}^i_{\text{train}}\}, \quad
\mathsf{N}^{\mathsf{f}}_j := \{f_j \notin \mathcal{F}_{\text{train}}\}, \quad
\mathsf{N}^{\mathsf{c}} := \{f_{\text{S}} \notin \mathcal{P}_{\text{train}}\},
\end{equation*}
indexed by atoms ($\mathsf{a}$), individual transformations ($\mathsf{f}$), and full compositions ($\mathsf{c}$). Assuming that failures induced by novel atoms, transformations, and transformation compositions contribute independently, the success probability factorizes as
\begin{equation}
\pi(C) \;=\; \pi_0 \prod_{i=0}^{l-1} \rho_{\mathsf{a}}^{\mathbb{1}[\mathsf{N}^{\mathsf{a}}_i]} \prod_{j=1}^{k} \rho_{\mathsf{f}}^{\mathbb{1}[\mathsf{N}^{\mathsf{f}}_j]} \, \rho_{\mathsf{c}}^{\mathbb{1}[\mathsf{N}^{\mathsf{c}}]},
\end{equation}
where $\pi_0 \in (0, 1]$ is the baseline in-distribution success probability and $\rho_{\mathsf{a}}, \rho_{\mathsf{f}}, \rho_{\mathsf{c}} \in (0, 1)$ are per-axis degradation factors. Defining the positive constants $\xi_{\mathsf{a}} := -\ln \rho_{\mathsf{a}}$, $\xi_{\mathsf{f}} := -\ln \rho_{\mathsf{f}}$, $\xi_{\mathsf{c}} := -\ln \rho_{\mathsf{c}}$, we obtain
\begin{equation}
\label{eq2proof_mid}
\ln \pi(C) \;=\; \ln \pi_0 - \xi_{\mathsf{a}} \sum_{i=0}^{l-1} \mathbb{1}[\mathsf{N}^{\mathsf{a}}_i] - \xi_{\mathsf{f}} \sum_{j=1}^{k} \mathbb{1}[\mathsf{N}^{\mathsf{f}}_j] - \xi_{\mathsf{c}}\, \mathbb{1}[\mathsf{N}^{\mathsf{c}}].
\end{equation}

\begin{lemma}[Task-complexity upper bound]
\label{lem:tgc_bound}
Let $\kappa := \min\!\left(\xi_{\mathsf{a}}/\alpha,\; \xi_{\mathsf{f}}/\beta,\; \xi_{\mathsf{c}}/\gamma\right) > 0$. Then
\begin{equation}
\label{eq2proof_final}
\ln \pi(C) \;\le\; \ln \pi_0 - \kappa \cdot \mathcal{T}(C).
\end{equation}
\end{lemma}

\begin{proof}[Proof of Lemma~\ref{lem:tgc_bound}]
By definition of $\kappa$, each term of \eqref{eq2proof_mid} dominates the corresponding $\kappa$-scaled term of \eqref{eq:tgc}:
$\xi_{\mathsf{a}} \sum_i \mathbb{1}[\mathsf{N}^{\mathsf{a}}_i] \ge \kappa\alpha \sum_i \mathbb{1}[\mathsf{N}^{\mathsf{a}}_i]$, and analogously for the $\xi_{\mathsf{f}}$ and $\xi_{\mathsf{c}}$ terms. Summing and negating yields \eqref{eq2proof_final}.
\end{proof}

Setting $\tau := (\ln \pi_0)/\kappa$, Lemma~\ref{lem:tgc_bound} gives, whenever $\mathcal{T}(C) > \tau$,
\begin{equation}
\ln \pi(C) \;\le\; \ln \pi_0 - \kappa \cdot \mathcal{T}(C) \;=\; -\kappa\bigl(\mathcal{T}(C) - \tau\bigr).
\end{equation}
Exponentiating both sides yields the claimed bound $\pi(C) \le \exp \bigl(-\kappa\,(\mathcal{T}(C) - \tau)\bigr)$.
\end{proof}

\subsection{Length Discrepancy Measure}
\label{app:length_distribution}
Length generalization examines how CoT reasoning degrades when models encounter test cases that differ in length from their training distribution. The difference in length could be introduced from the text space or the reasoning space of the problem. Therefore, we decompose length generalization into two complementary aspects: text length generalization and reasoning step generalization. Guided by intuition, we first propose to measure the length discrepancy.

\begin{proposition}[Squared-Exponential Length-Extrapolation Decay]
\label{prop:length_gaussian}
For a model trained on chain-of-thought sequences of fixed length $L_{\text{train}}$, the generalization error at test length $L$ admits a squared-exponential form in the length gap $|L - L_{\text{train}}|$:
\begin{equation}
\varepsilon(L) \;=\; \varepsilon_0 + (1 - \varepsilon_0)\left(1 - \exp\!\left(-\frac{(L - L_{\text{train}})^2}{2\sigma^2}\right)\right),
\end{equation}
where $\varepsilon_0 \in [0, 1]$ is the in-distribution error at $L = L_{\text{train}}$, $\sigma > 0$ is a length-generalization width parameter, and $L$ is the test sequence length. This form is a modeling ansatz motivated empirically; a heuristic derivation is given below.
\end{proposition}

\label{app:proof_length_bound}
\paragraph{Heuristic derivation.}
The following argument motivates the squared-exponential form stated in Proposition~\ref{prop:length_gaussian}; a fully rigorous derivation that tracks the internal dynamics of a transformer is beyond the scope of this work.

Consider a model $f_\theta$ trained on sequences of length $L_{\text{train}}$. For a test length $L \ne L_{\text{train}}$, the input induces a hidden-state distribution that differs from the training distribution in two respects: (i) positional encodings for positions $i > L_{\text{train}}$ have never been observed when $L > L_{\text{train}}$, and (ii) the learned attention patterns are calibrated for length $L_{\text{train}}$.

Let $q_{\text{test}}$ and $q_{\text{train}}$ denote the hidden-state distributions induced by test and training lengths, respectively. Under the mild assumption that per-position deviations accumulate independently with bounded variance, a second-order expansion of the log-likelihood of $q_{\text{test}}$ around $L = L_{\text{train}}$ produces a divergence of the form
\begin{equation}
D_{\mathrm{KL}}\bigl(q_{\text{test}} \,\|\, q_{\text{train}}\bigr) \;=\; \Theta\!\left((L - L_{\text{train}})^2\right),
\end{equation}
where the linear term vanishes because the model is trained to match the distribution at $L = L_{\text{train}}$, so the likelihood is stationary there.

Decomposing the test-length error as $\varepsilon(L) = \varepsilon_0 + \varepsilon_{\text{shift}}(L)$, with $\varepsilon_0$ the in-distribution error and $\varepsilon_{\text{shift}}(L_{\text{train}}) = 0$, we model $\varepsilon_{\text{shift}}$ as a transition from $0$ to its saturating value $1 - \varepsilon_0$ governed by the above divergence. A natural functional choice that matches this behavior is
\begin{equation}
\varepsilon_{\text{shift}}(L) \;=\; (1 - \varepsilon_0)\left(1 - \exp\!\left(-\frac{(L - L_{\text{train}})^2}{2\sigma^2}\right)\right),
\end{equation}
where the squared-exponential kernel reflects the quadratic growth of $D_{\mathrm{KL}}$ in the length gap. Substituting into $\varepsilon(L) = \varepsilon_0 + \varepsilon_{\text{shift}}(L)$ yields the expression in Proposition~\ref{prop:length_gaussian}.

This form satisfies three sanity checks: (a) $\varepsilon(L_{\text{train}}) = \varepsilon_0$, so there is no excess error at the training length; (b) $\lim_{|L - L_{\text{train}}| \to \infty} \varepsilon(L) = 1$, so the error remains bounded by $1$ regardless of the length gap; and (c) the curve is symmetric in $|L - L_{\text{train}}|$, consistent with the empirical curves in Fig.~\ref{fig:text_length}. The width parameter $\sigma$ is treated as an empirical fit parameter rather than a quantity derived from first principles.

The length-axis discrepancy is obtained from the excess error $\varepsilon(L) - \varepsilon_0$ through a monotone mapping,
\begin{equation}
\label{eq:delta_length}
\Delta_{\text{length}} \;=\; \psi_{\text{length}}\bigl(\varepsilon(L) - \varepsilon_0\bigr),
\end{equation}
where $\psi_{\text{length}} : [0,\, 1 - \varepsilon_0] \to [0, 1]$ is monotonically non-decreasing and satisfies $\psi_{\text{length}}(0) = 0$, so that $\Delta_{\text{length}}$ vanishes at $L = L_{\text{train}}$ and grows monotonically in $|L - L_{\text{train}}|$. The identity choice $\psi_{\text{length}}(x) = x$ recovers the raw excess error $\varepsilon(L) - \varepsilon_0$ given by Proposition~\ref{prop:length_gaussian}. Through this mapping, $\Delta_{\text{length}}$ plugs directly into the distribution-discrepancy term of Theorem~\ref{thm:generalization_bound}.

\subsection{Format Discrepancy Measure}
\label{app:format_distribution}
Format generalization assesses the robustness of CoT reasoning to surface-level variations in test queries. This dimension is especially crucial for determining whether models have internalized flexible, transferable reasoning strategies or remain reliant on the specific templates and phrasings encountered during training. We introduce a metric for measuring prompt similarity:

\begin{definition}[Format Alignment Score]
\label{def:fas}
For a training prompt distribution $\Pi_{\text{train}}$ and a test prompt $p_{\text{test}}$, we define the format alignment score
\begin{equation}
\mathcal{S}(p_{\text{test}}) \;:=\; \max_{p \in \Pi_{\text{train}}} \cos\bigl(\eta(p),\, \eta(p_{\text{test}})\bigr),
\end{equation}
where $\eta(\cdot)$ is a prompt embedding function. The format-axis discrepancy is obtained from the alignment gap $1 - \mathcal{S}(p_{\text{test}})$ through a monotone mapping,
\begin{equation}
\label{eq:delta_format}
\Delta_{\text{format}} \;=\; \psi_{\text{format}}\bigl(1 - \mathcal{S}(p_{\text{test}})\bigr),
\end{equation}
where $\psi_{\text{format}} : [0, 1] \to [0, 1]$ is monotonically non-decreasing and satisfies $\psi_{\text{format}}(0) = 0$ (perfect alignment $\Rightarrow$ zero format-axis discrepancy). The identity choice $\psi_{\text{format}}(x) = x$ recovers $1 - \mathcal{S}(p_{\text{test}})$. Through this mapping, $\Delta_{\text{format}}$ plugs directly into the distribution-discrepancy term of Theorem~\ref{thm:generalization_bound}.
\end{definition}

The four perturbation modes introduced in Section~\ref{sec:format_generalization} affect $\mathcal{S}$ in distinct ways. \emph{Insert} introduces a novel \texttt{<noise>} token, which shifts the overall embedding by a magnitude determined by $\phi$'s sensitivity to OOV symbols; \emph{Modify} replaces an in-vocabulary token with a novel one, producing a similar shift but without changing sequence length; \emph{Delete} removes an in-vocabulary token, shortening the prompt and redistributing attention mass within $\phi$; and \emph{Hybrid} combines all three, compounding their individual effects. In all four cases, $\mathcal{S}$ is expected to be monotonically non-increasing in the noise level $p$, so that larger $p$ induces larger $\Delta_{\text{format}}$ and, through Theorem~\ref{thm:generalization_bound}, a larger upper bound on test risk --- consistent with the degradation curves observed in Figure~\ref{fig:format_generalization}.

\begin{table*}[!htbp]
\centering
\caption{Evaluation on transformation generalization.}
\label{tab:novel_transformation}
\resizebox{0.95\linewidth}{!}{%
\begin{tabular}{@{}l|c|c|c|c|c|c|c|c|c@{}}
\toprule
\multirow{2.8}{*}{\centering\textbf{Transformation (Train $\rightarrow$ Test)}}
& \multicolumn{3}{c|}{\textbf{Exact Match (\%)}}
& \multicolumn{3}{c|}{\textbf{Edit Distance}}
& \multicolumn{3}{c}{\textbf{BLEU Score}} \\
\cmidrule(lr){2-4} \cmidrule(lr){5-7} \cmidrule(lr){8-10}
& Reasoning & Answer & Full Chain
& Reasoning & Answer & Full Chain
& Reasoning & Answer & Full Chain \\
\midrule
$\{f_2 \circ f_3,\; f_3 \circ f_2,\; f_3 \circ f_3\} \rightarrow f_2 \circ f_2$
& 6.66 & 10.25 & 6.66
& 0.0718 & 0.2244 & 0.0941
& 0.6683 & 0.1982 & 0.5417 \\

$\{f_2 \circ f_3,\; f_3 \circ f_2,\; f_2 \circ f_2\} \rightarrow f_3 \circ f_3$
& 100.00 & 9.19 & 9.19
& 0.0000 & 0.1768 & 0.0488
& 1.0000 & 0.1932 & 0.8220 \\

$f_2 \circ f_3 \rightarrow f_3 \circ f_2$
& 0.00 & 0.00 & 0.00
& 0.3728 & 0.4808 & 0.2997
& 0.0019 & 0.0000 & 0.2000 \\

$f_3 \circ f_2 \rightarrow f_2 \circ f_3$
& 0.00 & 0.00 & 0.00
& 0.2249 & 0.4808 & 0.2334
& 0.0952 & 0.0000 & 0.2548 \\
\bottomrule
\end{tabular}%
}
\end{table*}

\subsection{Proof of the CoT Generalization Bound}
\label{app:gen_bound}
\begin{proof}[Proof of Theorem~\ref{thm:generalization_bound}]
We use the notation of Theorem~\ref{thm:generalization_bound}. Decomposing the gap between test and empirical training risk,
\begin{equation}
\label{eq:app-decomp}
R_{\text{test}}(f_\theta) - \hat R_{\text{train}}(f_\theta) \;=\; \underbrace{R_{\text{test}}(f_\theta) - R_{\text{train}}(f_\theta)}_{\text{(i) distribution-shift term}} \;+\; \underbrace{R_{\text{train}}(f_\theta) - \hat R_{\text{train}}(f_\theta)}_{\text{(ii) sampling term}},
\end{equation}
we bound the two terms separately.

By the variational characterization of total-variation distance, for any measurable function $g$ with $|g| \le B$,
\begin{equation}
\left|\mathbb{E}_{\mathcal{D}_{\text{train}}}[g] - \mathbb{E}_{\mathcal{D}_{\text{test}}}[g]\right| \;\le\; 2B \cdot \operatorname{TV}(\mathcal{D}_{\text{train}}, \mathcal{D}_{\text{test}}).
\end{equation}
Taking $g(x, y) = \ell(f_\theta(x), y)$, which lies in $[0, B]$ by assumption, yields
\begin{equation}
\label{eq:app-train-test}
R_{\text{test}}(f_\theta) \;\le\; R_{\text{train}}(f_\theta) + 2B\,\Delta(\mathcal{D}_{\text{train}}, \mathcal{D}_{\text{test}}).
\end{equation}

\paragraph{Sampling term.}
Let $Z_i := \ell(f_\theta(x_i), y_i) \in [0, B]$; the $Z_i$ are i.i.d.\ with mean $R_{\text{train}}(f_\theta)$ and empirical mean $\hat R_{\text{train}}(f_\theta)$. Hoeffding's inequality gives, for any $\varepsilon > 0$,
\begin{equation}
\Pr\!\left(R_{\text{train}}(f_\theta) - \hat R_{\text{train}}(f_\theta) \ge \varepsilon\right) \;\le\; \exp\!\left(-\frac{2n\varepsilon^2}{B^2}\right).
\end{equation}
Setting the right-hand side to $\delta$ and solving for $\varepsilon$, we obtain, with probability at least $1 - \delta$,
\begin{equation}
\label{eq:app-train-emp}
R_{\text{train}}(f_\theta) \;\le\; \hat R_{\text{train}}(f_\theta) + B\sqrt{\frac{\log(1/\delta)}{2n}}.
\end{equation}

\paragraph{Combining the two bounds.}
Substituting \eqref{eq:app-train-emp} into \eqref{eq:app-train-test}, with probability at least $1 - \delta$,
\begin{equation}
R_{\text{test}}(f_\theta) \;\le\; \hat R_{\text{train}}(f_\theta) + 2B\,\Delta(\mathcal{D}_{\text{train}}, \mathcal{D}_{\text{test}}) + B\sqrt{\frac{\log(1/\delta)}{2n}},
\end{equation}
which is the bound stated in Theorem~\ref{thm:generalization_bound}.
\end{proof}

\section{Additional Quantitative Results}
\label{app:quantitative}
This appendix extends the quantitative evaluation beyond the figures in the main paper, drilling into each generalization dimension (transformation, element, text length, format), the roles of temperature and model scale, and the internal and external validity of our findings.

\subsection{Transformation Generalization}
\label{app:quantitative:transformation}
\subsubsection{Detailed analysis.}

\paragraph{Aggregate metrics across ID, CMP, POOD, and OOD.}
For the instance shown in Table~\ref{tab:transformation_full_chain}, moving from in-distribution to composition, POOD, and OOD, exact-match accuracy drops from $1$ to $0.01$, $0$, and $0$, while edit distance increases from $0$ to $0.13$, $0.17$, and $0.30$ when tested on data with transformation $f_1 \circ f_1$. Apart from ID, LLMs fail to produce a correct full chain in almost all cases; they occasionally produce correct CoT reasoning under some composition and POOD conditions, but only accidentally.

\paragraph{Coincidental correctness: the \texttt{ANAN} case.}
As shown in Table~\ref{tab:transformation_performance}, when generalizing from $f_1 \circ f_2$ to $f_2 \circ f_2$ the model correctly answers $0.1\%$ of questions. A close examination reveals that this is a coincidence: for the query element $\texttt{ANAN}$, the two operations happen to produce the same output, an artifact examined in Appendix~\ref{app:qualitative}. Once such degenerate inputs are excluded, the $0.1\%$ residual disappears.

\paragraph{Commutativity-induced correctness with unfaithful reasoning.}
Breaking the full chain into reasoning steps and final answer reveals a systematic pattern. Under the composition-generalization setting, the reasoning steps are entirely correct on test transformations $f_1 \circ f_1$ and $f_2 \circ f_2$ yet the final answers are wrong: when a novel transformation (say $f_1 \circ f_1$) is present, the LLM generalizes by copying the most similar reasoning path seen in training ($f_1 \circ f_2$), producing a locally plausible trace attached to the wrong answer. Conversely, when generalizing from $f_1 \circ f_2$ to $f_2 \circ f_1$ (or vice versa), commutativity of the two orthogonal transformations yields \emph{correct} answers paired with \emph{unfaithful} reasoning paths. Collectively, these results show that CoT reasoning fails to generalize to novel transformations---even to novel compositions of well-learned primitives. Rather than demonstrating a true understanding of the task, CoT reasoning under task shifts appears to replicate patterns learned from training.

\subsubsection{Introducing another transformation.}
Our main conclusion is that CoT reasoning cannot generalize to genuinely novel transformations, including unseen compositions, even when the underlying primitives are well learned. Counterexamples based on commutativity have already substantiated this claim (Table~\ref{tab:transformation_performance}). We now extend the analysis by introducing a non-commutative transformation $f_3$ and evaluating generalization behaviors that cannot be trivially explained by commutative equivalence. The results are summarized in Table~\ref{tab:novel_transformation}.

When models are trained on mixtures of transformations involving $f_2$ and $f_3$ and evaluated on unseen compositions (e.g., ${f_2 \circ f_3, f_3 \circ f_2, f_3 \circ f_3} \rightarrow f_2 \circ f_2$), performance remains extremely poor across all metrics. Exact match accuracy for the full chain stays below 10\%, and both edit distance and BLEU scores indicate substantial divergence from the correct reasoning traces and final answers. Notably, even when the reasoning component occasionally achieves high exact match (e.g., 100\% reasoning accuracy in ${f_2 \circ f_3, f_3 \circ f_2, f_2 \circ f_2} \rightarrow f_3 \circ f_3$), the corresponding full-chain and answer-level accuracy collapse, revealing a clear disconnect between locally plausible reasoning steps and globally correct execution.

\label{app:quantitative:element}
\begin{wrapfigure}{!ht}{0.45\textwidth}
    \centering
    \centering
    \includegraphics[width=1\linewidth]{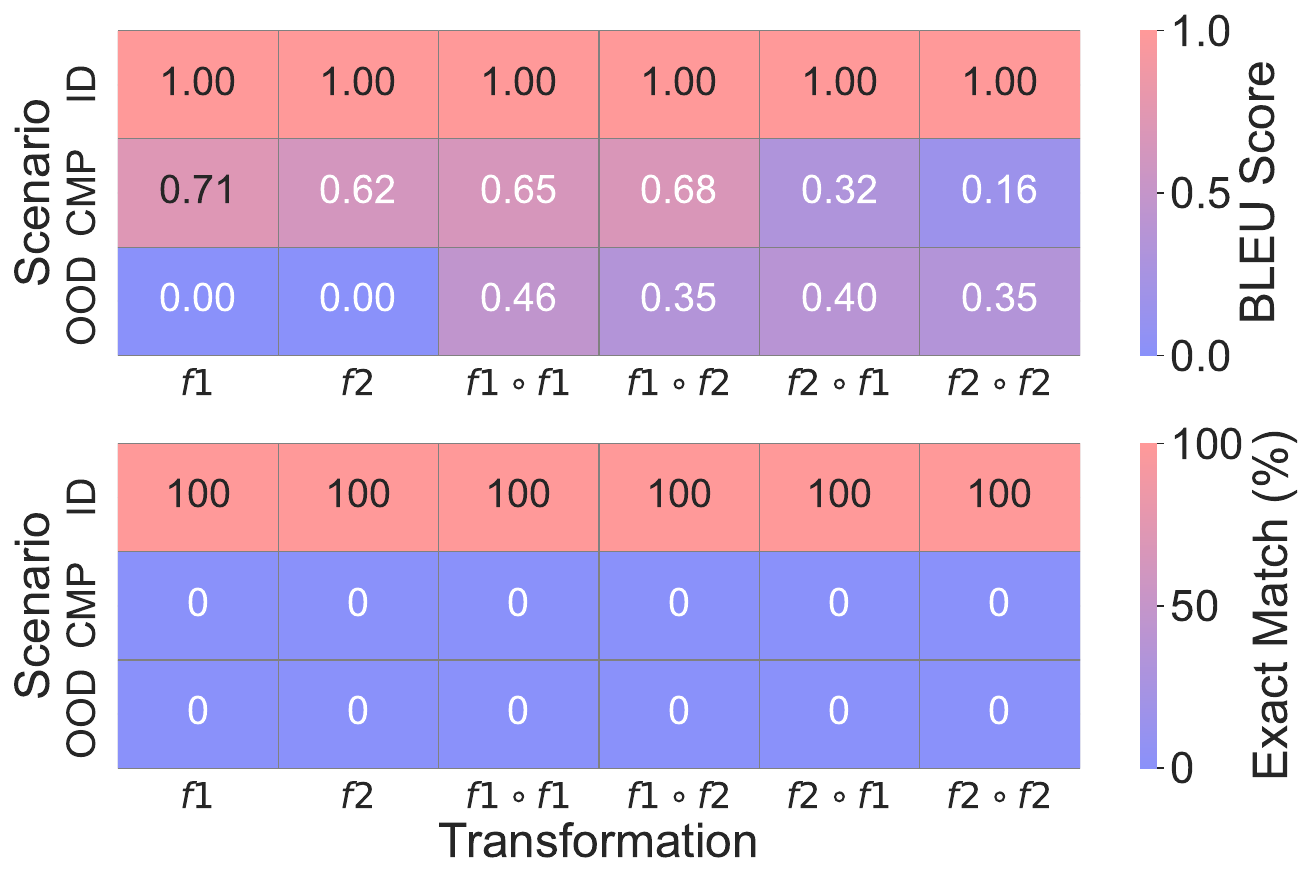}
    \caption{Element generalization results on various scenarios and relations.}
    \label{fig:element_generlization}
\end{wrapfigure}

More strikingly, in strictly non-commutative transfer settings such as $f_2 \circ f_3 \rightarrow f_3 \circ f_2$ and its reverse, the model fails completely: exact match drops to 0 across reasoning, answer, and full chain, while edit distance sharply increases and BLEU scores approach zero. Unlike earlier commutativity-induced cases---where incorrect reasoning paths could still yield correct answers---these failures demonstrate that once superficial equivalences are removed, CoT reasoning no longer exhibits any meaningful transfer. This provides strong evidence that prior apparent generalization was not driven by learning transformation semantics, but rather by exploiting distributional artifacts such as commutativity.

Overall, Table~\ref{tab:novel_transformation} reinforces our central claim: CoT reasoning does not support systematic generalization to novel transformations. Instead, its success hinges on structural overlap and distributional shortcuts present in the training data. When these shortcuts are eliminated via non-commutative transformations, both reasoning traces and answers degrade simultaneously, exposing the brittleness of CoT reasoning under genuine task-level distribution shifts. Illustrative examples of $f_3$ are provided in Appendix~\ref{app:dataalchemy} for completeness.

\subsection{Element Generalization}
Element generalization is another critical factor to consider when LLMs try to generalize to new tasks.

\noindent\textbf{Experiment settings.} Similar to transformation generalization, we fix other factors and consider three progressive distribution shifts for elements: ID, CMP, and OOD, as shown in Figure~\ref{fig:main}. It is noted that in composition, we test if CoT reasoning can be generalized to novel combinations when seeing all the basic atoms in the elements, e.g., $(\texttt{A}, \texttt{B}, \texttt{C}, \texttt{D}) \rightarrow (\texttt{B}, \texttt{C}, \texttt{D}, \texttt{A})$. Based on the atom order in combination (can be measured by edit distance $n$), the CMP can be further developed. While for OOD, atoms that constitute the elements are totally unseen during the training.

\begin{figure}[!ht]
\centering
\graphicspath{{figs/}}

\subfloat[Performance on unseen element via SFT in various CMP scenarios.]{%
  \includegraphics[width=0.49\linewidth]{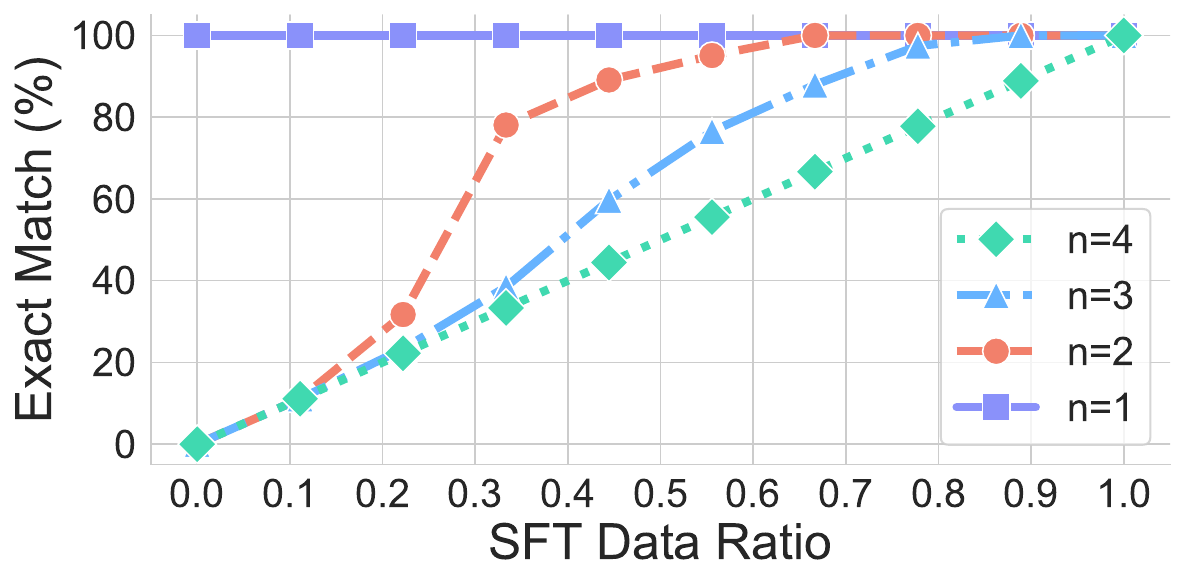}%
  \label{fig:element_sft}
}\hfill
\subfloat[Evaluation of CoT reasoning in SFT.]{%
  \includegraphics[width=0.49\linewidth]{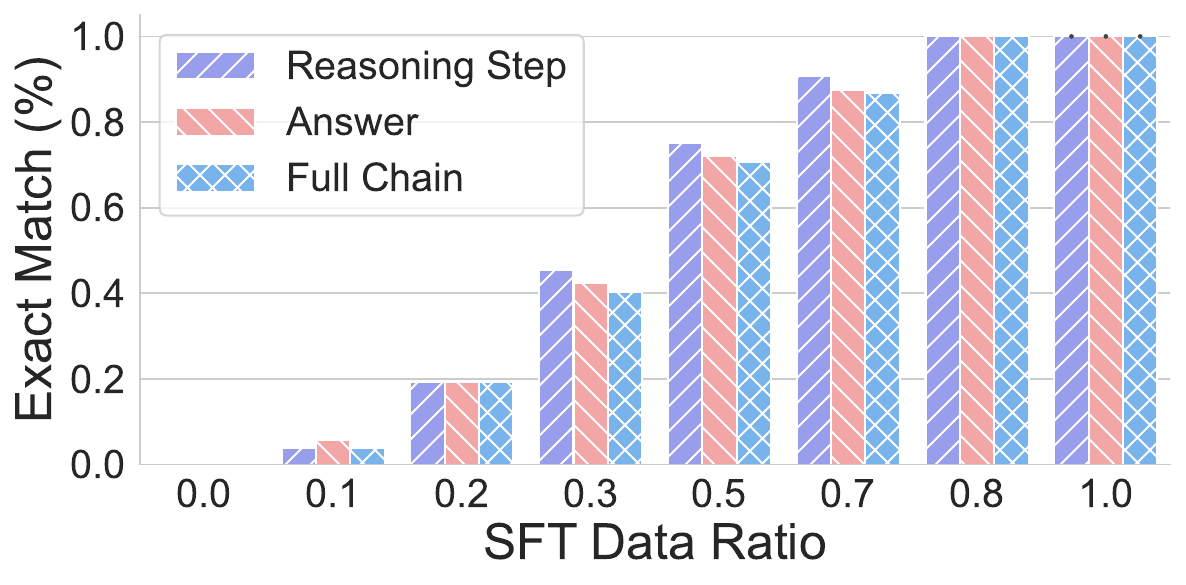}%
  \label{fig:element_sft_details}
}

\caption{SFT performances for element generalization. SFT helps to generalize to novel elements.}
\label{fig:element_generalization_sft}
\end{figure}

\noindent\textbf{Findings.}

\paragraph{Aggregate performance across ID/CMP/OOD.}
Similar to transformation generalization, performance degrades sharply under element-distribution shift across all transformations, as shown in Figure~\ref{fig:element_generlization}. From ID to CMP and OOD, exact match drops from $1.0$ to $0$ in every case; most strikingly, BLEU score also collapses to $0$ when transferred to $f_1$ and $f_2$. A failure case in Appendix~\ref{app:qualitative} shows that models fail to produce any sensible output when the test elements contain novel atoms.

\paragraph{SFT recovery curve.}
We further explore when CoT reasoning can generalize to novel elements by conducting SFT; results are summarised in Figure~\ref{fig:element_generalization_sft}. We evaluate full-chain exact match under three CMP scenarios parameterized by the edit distance $n$ between training and test elements. Mirroring the SFT-on-transformation results, performance recovers rapidly when the SFT set contains examples close to the test set (small $n$). Notably, the exact-match rate plateaus at the lower-bound performance once $n = 3$, suggesting that CoT reasoning generalizes only very locally to novel elements even after targeted SFT.

\paragraph{Reasoning/answer mismatch during SFT.}
Analyzing reasoning, answer, and token-level exact match during training for $n = 3$ (Figure~\ref{fig:element_sft_details}), we observe a mismatch between answer accuracy and reasoning-step accuracy across the SFT trajectory. The model can learn to produce the correct final answer without the reasoning steps catching up, and vice versa. This decoupling offers a mechanistic hint for the inconsistency between traces and answers reported in the main results.

\begin{table*}[!htbp]
\centering
\caption{Evaluation on text length generalization.}
\label{tab:text_length}
\resizebox{0.8\linewidth}{!}{%
\begin{tabular}{@{}c|c|c|c|c|c|c|c|c|c@{}}
\toprule
\multirow{2.8}{*}{\centering\textbf{Text Length}}
& \multicolumn{3}{c|}{\textbf{Exact Match (\%)}}
& \multicolumn{3}{c|}{\textbf{Edit Distance}}
& \multicolumn{3}{c}{\textbf{BLEU Score}} \\
\cmidrule(lr){2-4} \cmidrule(lr){5-7} \cmidrule(lr){8-10}
& Reasoning & Answer & Full Chain
& Reasoning & Answer & Full Chain
& Reasoning & Answer & Full Chain \\
\midrule
2 & 0.00 & 0.00 & 0.00 & 0.4969 & 0.5000 & 0.3772 & 0.1186 & 0.0000 & 0.4214 \\
3 & 0.00 & 0.00 & 0.00 & 0.3203 & 0.2540 & 0.2221 & 0.1519 & 0.0000 & 0.5471 \\
4 & 100.00 & 100.00 & 100.00 & 0.0000 & 0.0000 & 0.0000 & 1.0000 & 1.0000 & 1.0000 \\
5 & 0.00 & 0.00 & 0.00 & 0.2667 & 0.2000 & 0.1818 & 0.1958 & 0.2688 & 0.6220 \\
6 & 0.00 & 0.00 & 0.00 & 0.4816 & 0.3337 & 0.3294 & 0.1174 & 0.2077 & 0.4763 \\
\bottomrule
\end{tabular}%
}
\end{table*}

\subsection{Text Length Generalization}
\label{app:quantitative:text_length}
Text length generalization evaluates how CoT performance varies when the input text length (i.e., the element length $l$) differs from training examples. Considering the way LLMs process long text, this aspect is crucial because real-world problems often involve varying degrees of complexity that manifest as differences in problem statement length, context size, or information density.

\begin{wrapfigure}{!ht}{0.45\textwidth}
    \centering
    \includegraphics[width=1\linewidth]{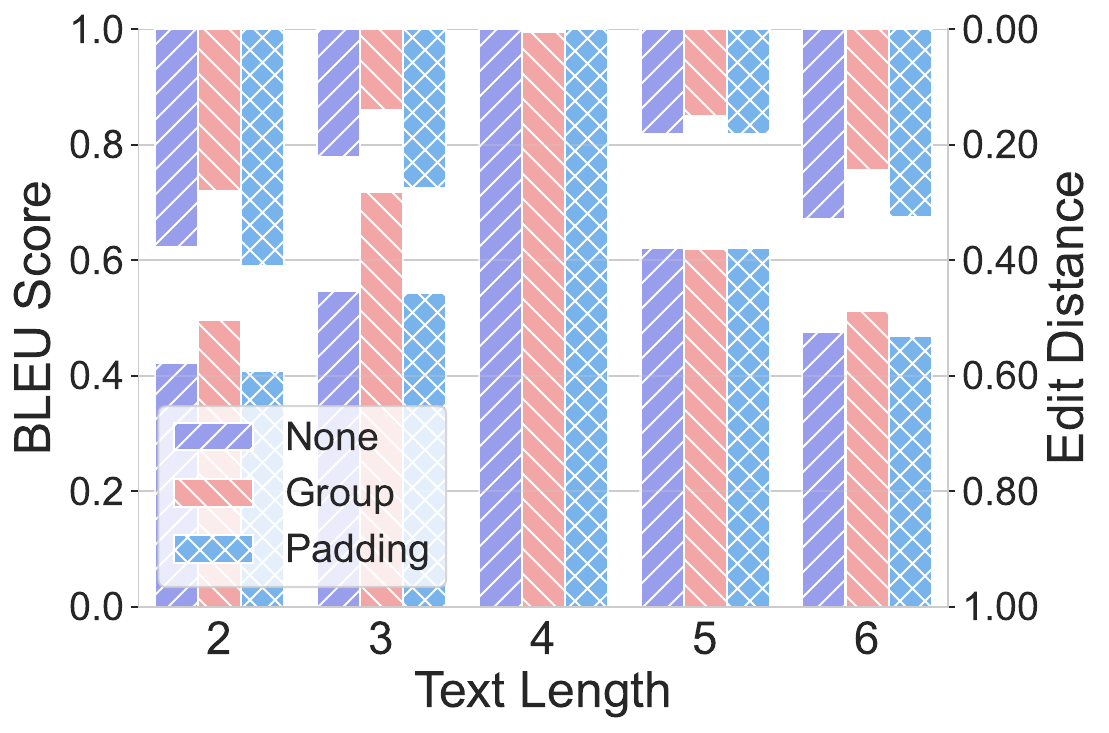}
    \caption{Performance of text length generalization across various padding strategies. Group strategies contribute to length generalization.}
    \label{fig:text_length_padding}
\end{wrapfigure}

\noindent\textbf{Experiment settings.} We pre-train LLMs on the dataset with text length merely on $l=4$ while fixing other factors and evaluate the performance on a variety of lengths. We consider three different padding strategies during the pre-training: (i) None: LLMs do not use any padding. (ii) Padding: We pad the LLM to the max length of the context window. (iii) Group: We group the text and truncate it into segments with a maximum length.

\noindent\textbf{Findings.} As illustrated in the Table~\ref{tab:text_length}, the CoT reasoning failed to directly generate two test cases even though those lengths present a mild distribution shift. Further, the performance declines as the length discrepancy increases, as shown in Figure~\ref{fig:text_length_padding}. For instance, from data with $l=4$ to those with $l=3$ or $l=5$, the BLEU score decreases from 1 to 0.55 and 0.62. Examples in Appendix~\ref{app:qualitative} indicate that LLMs attempt to produce CoT reasoning with the same length as the training data by adding or removing tokens in the reasoning chains. The efficacy of CoT reasoning length generalization deteriorates as the discrepancy increases. Moreover, we consider using a different padding strategy to decrease the divergence between the training data and test cases. We found that padding to the max length does not contribute to length generalization. However, the performance increases when we replace the padding with text by using the group strategy. The mechanism is simple: fixed-length padding keeps every training input at exactly the same length, so the model never sees a distribution over lengths during pre-training; the group strategy, in contrast, exposes the model to a distribution of effective text lengths by grouping and truncating sequences, thereby widening the training distribution along the length axis and making mild length shifts at test time look in-distribution.

\subsection{Format Generalization}
\label{app:quantitative:format}
Format generalization quantifies how surface-level perturbations of the prompt degrade CoT reasoning. Complementing the main-paper treatment in Section~\ref{sec:format_generalization}, this subsection consolidates the per-mode quantitative behavior.

\noindent\textbf{Experiment settings.} We apply the four perturbation modes defined in Section~\ref{sec:format_generalization} --- \emph{Insert}, \emph{Delete}, \emph{Modify}, and \emph{Hybrid} --- at noise levels $p \in \{5\%, 10\%, 15\%, 20\%, 25\%, 30\%\}$ to the test queries of the in-distribution setting, so that any degradation is attributable solely to the format shift.

\noindent\textbf{Findings.} Figure~\ref{fig:format_generalization} shows that all four perturbation modes produce smooth, monotone degradation curves, with \emph{Insert} typically inflicting the largest drop at any fixed $p$ because it injects an entirely OOV token that shifts every downstream attention pattern. \emph{Delete} and \emph{Modify} cause comparable but smaller drops, while \emph{Hybrid} --- composing all three --- roughly tracks the envelope of the individual modes. Table~\ref{tab:external_format_generalization} confirms the same trend for SOTA LLMs: for both LLaMA3-8B-Instruct and Qwen3-14B-Instruct, exact-match accuracy decays from $100\%$ at $p=0\%$ to near-zero at $p=30\%$, with edit distance and BLEU score degrading in lockstep. This mirrors the monotonicity predicted by $\mathcal{S}$ in Appendix~\ref{app:format_distribution}: larger $p$ reduces $\mathcal{S}(p_{\text{test}})$, which in turn loosens the bound of Theorem~\ref{thm:generalization_bound}.

\subsection{Temperature and Model Size}
\label{app:quantitative:model_size}
Temperature and model size generalization explores how variations in sampling temperature and model capacity can influence the stability and robustness of CoT reasoning. For the sake of rigorous evaluation, we further investigate whether different choices of temperatures and model sizes may significantly affect our results.

\noindent\textbf{Experiment settings.} We explore the impact of different temperatures on the validity of the presented results. We adopt the same setting in the transformation generalization.

\begin{figure}[!th]
\centering
\graphicspath{{figs/}}

\subfloat[Influences of various temperatures.]{%
  \includegraphics[width=0.42\linewidth]{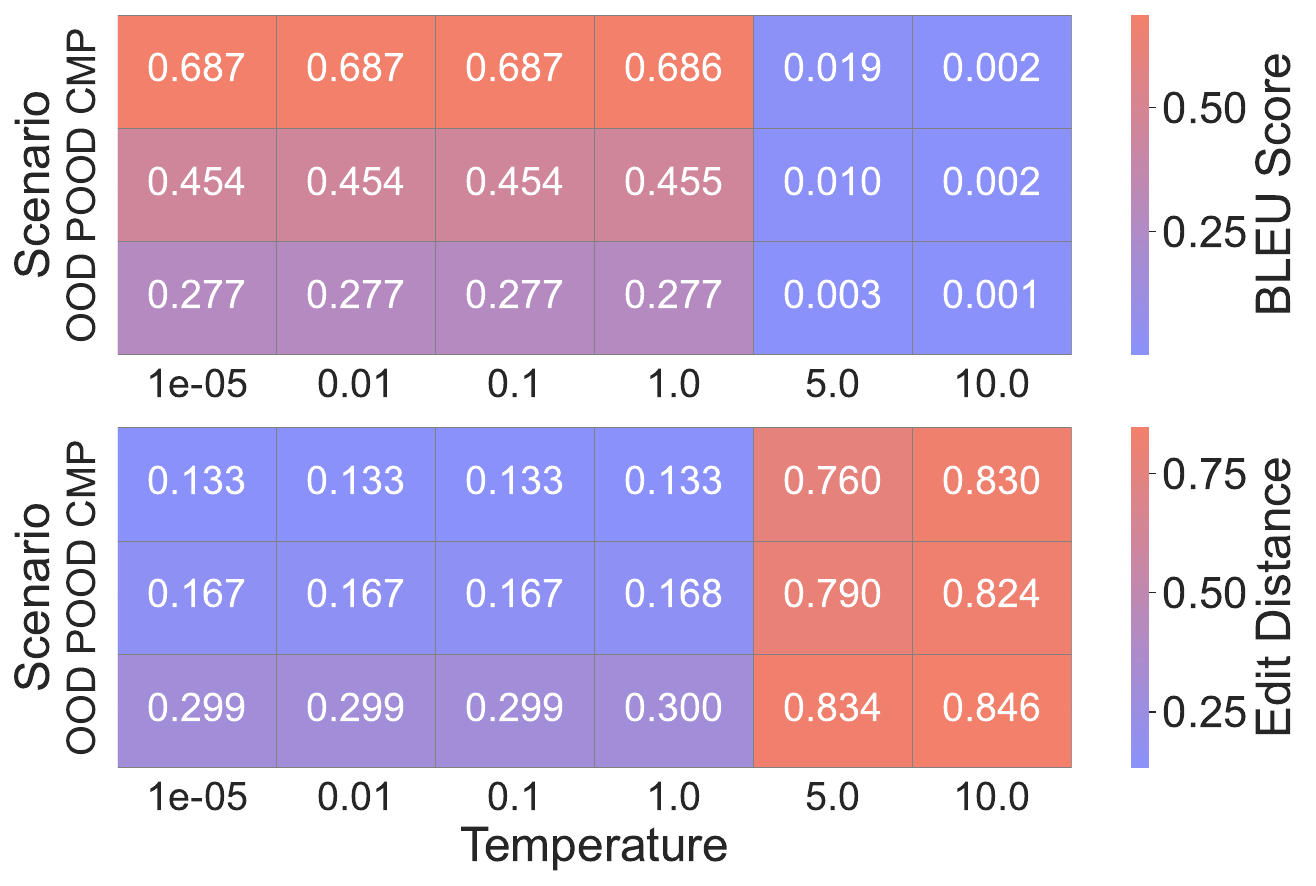}%
  \label{fig:temperature}
}\hspace{8mm}
\subfloat[Influences of various sizes.]{%
  \includegraphics[width=0.42\linewidth]{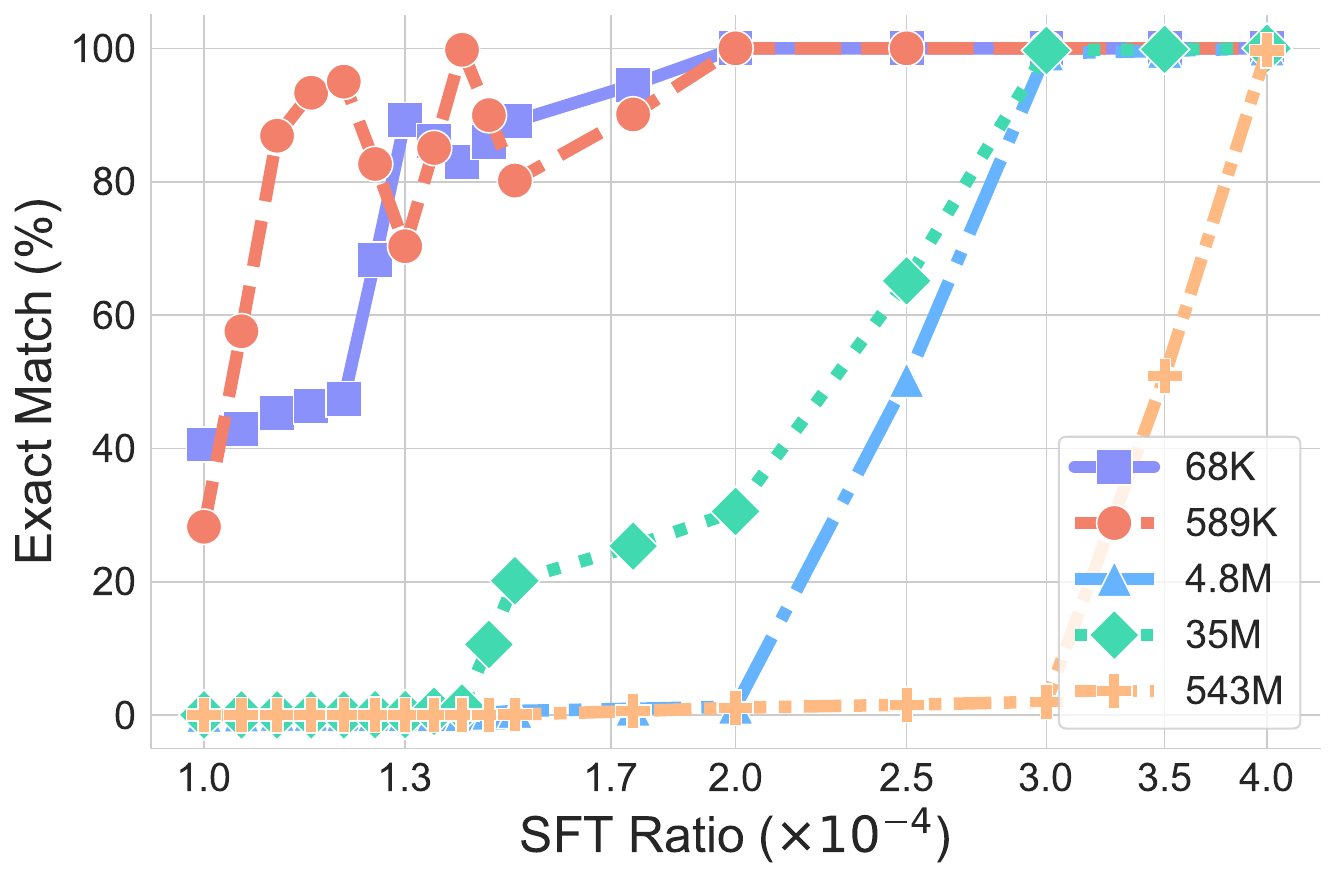}%
  \label{fig:model_size}
}

\caption{Temperature and model size. The findings hold under different temperatures and model sizes.}
\label{fig:temp_size}
\end{figure}

\noindent\textbf{Findings.} As illustrated in Figure~\ref{fig:temperature}, LLMs tend to generate consistent and reliable CoT reasoning across a broad range of temperature settings (from $10^{-5}$ up to $1$, with higher values up to $10$ tested but excluded from the figure because decoding becomes essentially uniform sampling), provided the temperature remains within a suitable range. This stability is maintained even when the models are evaluated under a variety of distribution shifts, indicating that the trends we report are not artefacts of a particular decoding configuration.

\noindent\textbf{Experiment settings.} We further examine the influence of model size by employing the same experimental configuration as used in the novel relation SFT study. In particular, we first pretrain models of different sizes using the transformation $f_1 \circ f_1$, and subsequently perform SFT on $f_2 \circ f_2$ while varying the SFT ratios.

\noindent\textbf{Finding.}
Fig.~\ref{fig:model_size} shows the accuracy of models with different sizes using different SFT ratios, which closely matches the result of our default model size across all evaluated settings and configurations. Notably, at any fixed SFT ratio, larger models reach near-perfect ID accuracy more quickly, but exhibit the same OOD collapse as smaller models once the SFT support is exhausted. This indicates that model scale accelerates interpolation within the SFT-expanded training distribution rather than enabling extrapolation beyond it --- a finding that reinforces the data-distribution lens at the level of model capacity.

\subsection{Internal Validity}
\label{app:quantitative:internal}
\begin{figure*}[!th]
    \centering
    \includegraphics[width=0.32\linewidth]{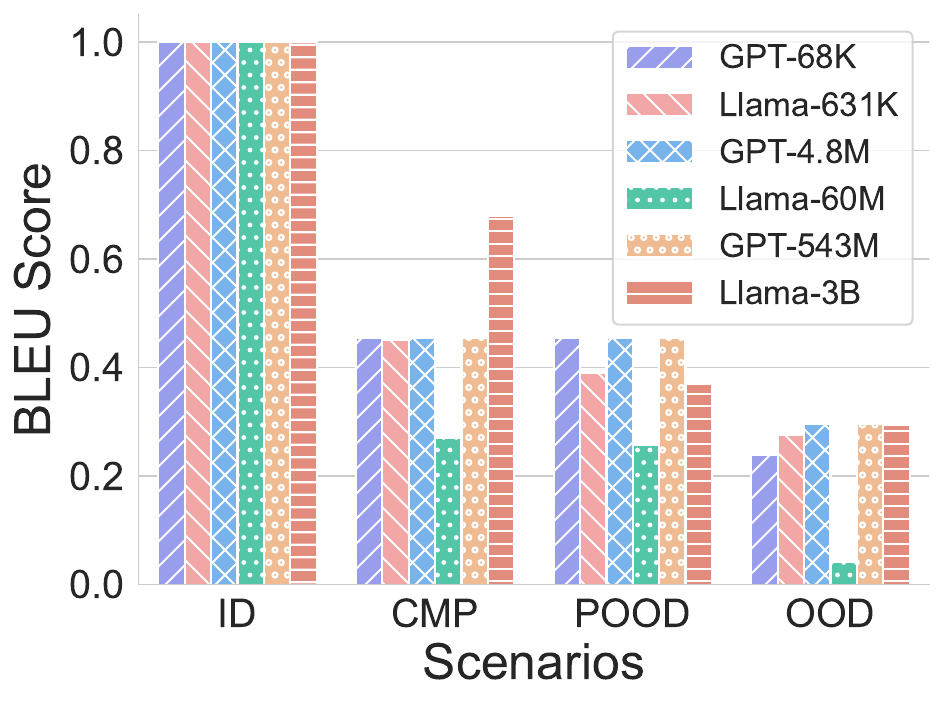}
    \includegraphics[width=0.32\linewidth]{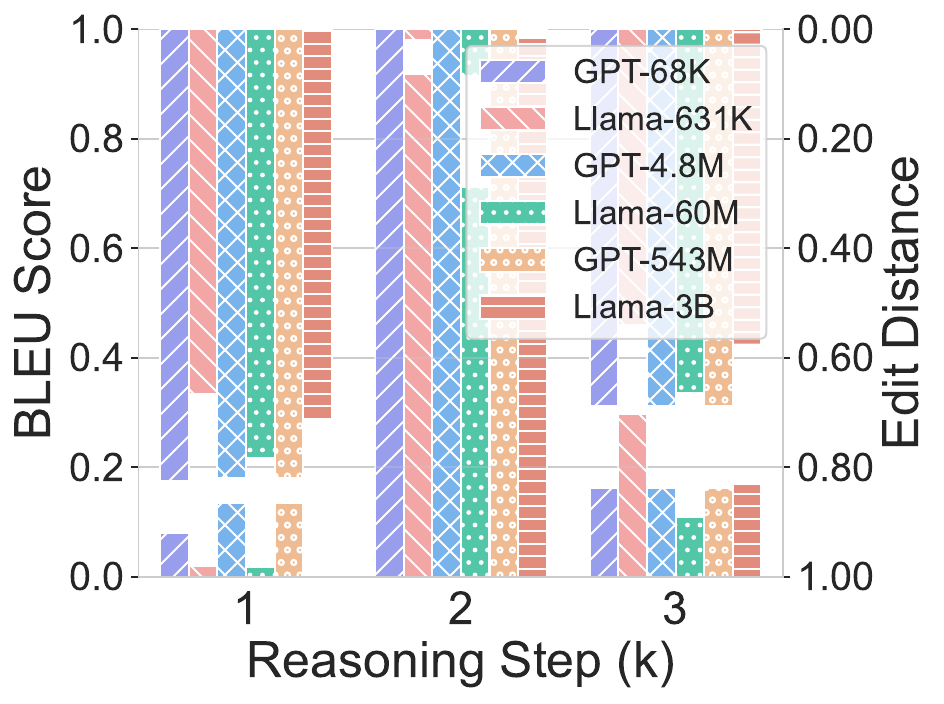}
    \includegraphics[width=0.32\linewidth]{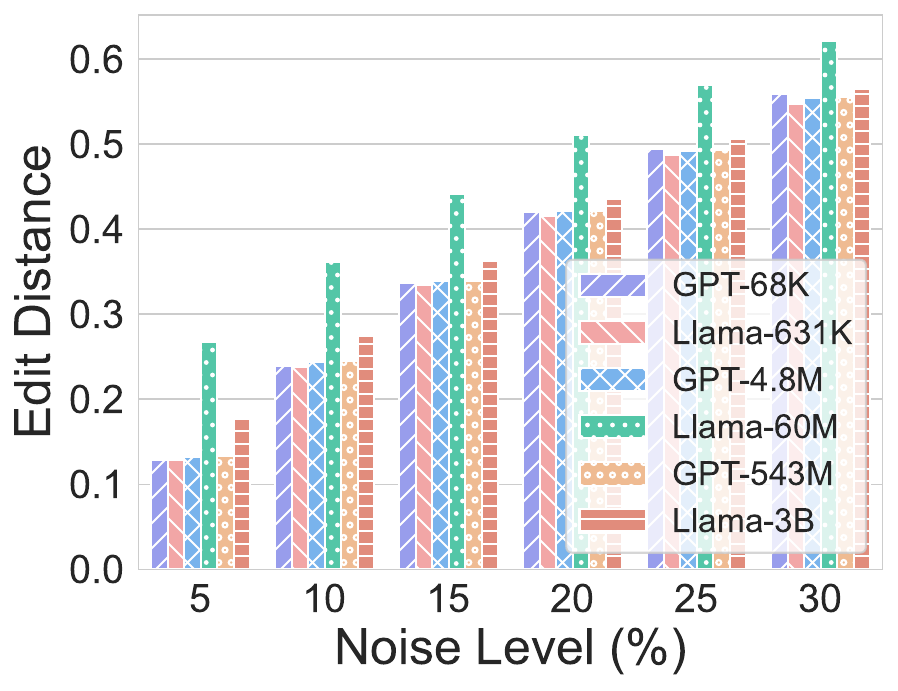}
    \caption{Task, length, and format generalization of LLMs with settings. The data distribution lens is invariant across LLMs with various sizes and architectures.}
    \label{fig:size_and_architecture_app}
\end{figure*}

Figures~\ref{fig:size_and_architecture} and~\ref{fig:size_and_architecture_app} illustrate task, length, and format generalization across a wide range of GPT- and LLaMA-style models. Increasing distribution discrepancy leads to a monotonic degradation of CoT performance regardless of architectural choice or parameter count. While larger models achieve uniformly higher absolute scores under near in-distribution conditions, they do not exhibit qualitatively different robustness profiles under moderate or severe shifts. This suggests that the observed failures of CoT reasoning cannot be attributed to insufficient capacity or architectural idiosyncrasies, but rather reflect a shared inductive bias learned from training distributions.

Model scaling consistently improves performance in ID and mildly shifted regimes (e.g., CMP or small reasoning-step extrapolation), but provides diminishing returns as the discrepancy increases toward POOD and OOD settings. In particular, larger models tend to preserve fluent intermediate reasoning traces even when final answers deteriorate, mirroring the same failure modes observed in smaller models. This pattern reinforces the interpretation that scaling primarily enhances pattern interpolation within the support of the training distribution, rather than enabling principled extrapolation beyond it.

Notably, the qualitative nature of errors remains stable across model sizes and architectures. Under task shifts, models consistently default to the closest seen transformation pattern; under length shifts, they bias toward producing training-length reasoning chains; and under format perturbations, they remain sensitive to surface-level noise in structurally salient regions. The persistence of these behaviors across settings further supports the internal validity of the data distribution lens: CoT reasoning behaves as a distribution-sensitive generative process rather than an architecture-specific reasoning mechanism.

Taken together, these results demonstrate that our core findings are not artifacts of a particular model family, scale, or training instability. Instead, the dependence of CoT effectiveness on task, length, and format distributions emerges as a stable and reproducible phenomenon across controlled LLM instantiations. This strengthens the claim that distribution discrepancy---rather than model choice---is the dominant factor governing when CoT reasoning succeeds or fails.

\subsection{External Validity}
\label{app:quantitative:external}
As illustrated in Table~\ref{tab:external_task_generalization},~\ref{tab:external_reasoning_step_generalization} \&~\ref{tab:external_format_generalization}, the results on LLaMA3-8B and Qwen3-14B-Instruct confirm that the behaviors identified in \methodName{} persist in real-world, pretrained LLMs despite unknown and opaque training distributions. In the in-distribution (ID) setting, both models achieve perfect performance (100\% exact match, zero edit distance, BLEU score of 1), indicating that the curated tasks are well within the expressive and optimization capacity of modern instruction-tuned models.

However, once distribution shifts are introduced, performance degrades sharply and systematically. Under composition (CMP), exact match accuracy drops to 8.52\% for LLaMA3-8B and near zero (0.01\%) for Qwen3-14B-Instruct, while POOD and OOD settings result in complete failure (0\% exact match) for both models. Correspondingly, edit distance increases and BLEU score decreases monotonically from CMP to OOD. These trends mirror those observed in models trained from scratch, reinforcing that the failure of CoT reasoning under task-level distribution shifts is not an artifact of synthetic training or limited model scale.

\begin{table*}[!ht]
\centering
\caption{Task generalization performance of SOTA LLMs (mean $\pm$ std).}
\label{tab:external_task_generalization}
\resizebox{0.7\linewidth}{!}{%
\begin{tabular}{l|l|c|c|c}
\toprule
\textbf{Model} & \textbf{Scenario}
& \textbf{Exact Match (\%)}
& \textbf{Edit Distance}
& \textbf{BLEU Score} \\
\midrule
\multirow{4}{*}{LLaMA3-8B}
& ID
& $100.00 \pm 0.00$
& $0.00 \pm 0.00$
& $1.00 \pm 0.00$ \\
& CMP
& $8.52 \pm 0.00$
& $0.23 \pm 0.00$
& $0.61 \pm 0.00$ \\
& POOD
& $0.00 \pm 0.01$
& $0.25 \pm 0.01$
& $0.46 \pm 0.00$ \\
& OOD
& $0.00 \pm 0.00$
& $0.27 \pm 0.01$
& $0.27 \pm 0.00$ \\
\midrule
\multirow{4}{*}{Qwen3-14B-Instruct}
& ID
& $100.00 \pm 0.00$
& $0.00 \pm 0.00$
& $1.00 \pm 0.00$ \\
& CMP
& $0.01 \pm 0.01$
& $0.17 \pm 0.02$
& $0.61 \pm 0.00$ \\
& POOD
& $0.00 \pm 0.00$
& $0.26 \pm 0.01$
& $0.42 \pm 0.00$ \\
& OOD
& $0.00 \pm 0.00$
& $0.38 \pm 0.01$
& $0.36 \pm 0.00$ \\
\bottomrule
\end{tabular}%
}
\end{table*}

Reasoning-step generalization further highlights the limited extrapolation ability of CoT reasoning in SOTA models. Both LLaMA3-8B and Qwen3-14B-Instruct achieve perfect performance at the in-distribution reasoning depth ($k=2$), but fail almost entirely at unseen depths. For $k=1$ and $k=3$, exact match accuracy collapses to 0\% in nearly all cases, accompanied by large edit distances and sharply reduced BLEU scores. Notably, Qwen3-14B-Instruct exhibits a marginal non-zero accuracy (0.26\%) at $k=3$, but this gain is unstable and negligible relative to the ID performance. This pattern indicates that even large, instruction-tuned models do not acquire a length-agnostic or algorithmic reasoning procedure, but instead internalize a narrowly scoped reasoning template tied to the training distribution.

\begin{table*}[!ht]
\centering
\caption{Reasoning step generalization of SOTA LLMs (mean $\pm$ std).}
\label{tab:external_reasoning_step_generalization}
\resizebox{0.7\linewidth}{!}{%
\begin{tabular}{l|c|c|c|c}
\toprule
\textbf{Model}
& \textbf{Reasoning Step ($k$)}
& \textbf{Exact Match (\%)}
& \textbf{Edit Distance}
& \textbf{BLEU Score} \\
\midrule
\multirow{3}{*}{LLaMA3-8B}
& 1
& $0.00 \pm 0.00$
& $0.75 \pm 0.01$
& $0.18 \pm 0.00$ \\
& 2
& $100.00 \pm 0.00$
& $0.00 \pm 0.00$
& $1.00 \pm 0.00$ \\
& 3
& $0.00 \pm 0.00$
& $0.54 \pm 0.01$
& $0.40 \pm 0.00$ \\
\midrule
\multirow{3}{*}{Qwen3-14B-Instruct}
& 1
& $0.00 \pm 0.00$
& $0.54 \pm 0.02$
& $0.35 \pm 0.00$ \\
& 2
& $100.00 \pm 0.00$
& $0.00 \pm 0.00$
& $1.00 \pm 0.00$ \\
& 3
& $0.26 \pm 0.05$
& $0.65 \pm 0.08$
& $0.20 \pm 0.00$ \\
\bottomrule
\end{tabular}%
}
\end{table*}

Format generalization experiments show consistent degradation trends across noise levels for both models. As noise increases from 0\% to 30\%, exact match accuracy decays smoothly from 100\% to near zero, while edit distance increases and BLEU score decreases monotonically. Although LLaMA3-8B initially appears more sensitive at low noise levels (e.g., 32.30\% exact match at 5\% noise versus 9.96\% for Qwen3-14B-Instruct), both models converge to similarly poor performance under higher noise. The near-parallel degradation curves suggest that robustness to surface-level perturbations is constrained by distributional alignment.

\begin{table*}[!ht]
\centering
\caption{Format generalization under different noise levels for SOTA LLMs (mean $\pm$ std).}
\label{tab:external_format_generalization}
\resizebox{0.7\linewidth}{!}{%
\begin{tabular}{l|c|c|c|c}
\toprule
\textbf{Model}
& \textbf{Noise Level (\%)}
& \textbf{Exact Match (\%)}
& \textbf{Edit Distance}
& \textbf{BLEU Score} \\
\midrule
\multirow{7}{*}{LLaMA3-8B}
& 0
& $100.00 \pm 0.00$
& $0.00 \pm 0.00$
& $1.00 \pm 0.00$ \\
& 5
& $32.30 \pm 0.47$
& $0.41 \pm 0.02$
& $0.04 \pm 0.00$ \\
& 10
& $17.74 \pm 0.38$
& $0.45 \pm 0.00$
& $0.04 \pm 0.00$ \\
& 15
& $9.27 \pm 0.29$
& $0.49 \pm 0.05$
& $0.04 \pm 0.00$ \\
& 20
& $4.72 \pm 0.21$
& $0.53 \pm 0.03$
& $0.03 \pm 0.00$ \\
& 25
& $2.32 \pm 0.15$
& $0.57 \pm 0.00$
& $0.03 \pm 0.00$ \\
& 30
& $1.09 \pm 0.10$
& $0.60 \pm 0.08$
& $0.03 \pm 0.00$ \\
\midrule
\multirow{7}{*}{Qwen3-14B-Instruct}
& 0
& $100.00 \pm 0.00$
& $0.00 \pm 0.00$
& $1.00 \pm 0.00$ \\
& 5
& $9.96 \pm 0.30$
& $0.22 \pm 0.03$
& $0.41 \pm 0.00$ \\
& 10
& $5.09 \pm 0.22$
& $0.35 \pm 0.01$
& $0.24 \pm 0.00$ \\
& 15
& $2.43 \pm 0.15$
& $0.44 \pm 0.02$
& $0.16 \pm 0.00$ \\
& 20
& $1.19 \pm 0.11$
& $0.52 \pm 0.05$
& $0.11 \pm 0.00$ \\
& 25
& $0.53 \pm 0.07$
& $0.57 \pm 0.04$
& $0.08 \pm 0.00$ \\
& 30
& $0.22 \pm 0.05$
& $0.62 \pm 0.02$
& $0.06 \pm 0.00$ \\
\bottomrule
\end{tabular}%
}
\end{table*}

Across all settings, the reported standard deviations are small relative to the absolute performance gaps between ID and shifted distributions. This indicates that the observed trends are stable across runs and not driven by sampling noise or stochastic decoding effects. In particular, zero or near-zero variance in ID settings confirms deterministic mastery of in-distribution patterns, while low variance under OOD conditions reflects consistently poor generalization rather than brittle or erratic behavior.

Overall, these results establish strong external validity for the proposed data distribution lens. Despite their scale, architectural sophistication, and instruction tuning, SOTA LLMs exhibit the same qualitative behaviors as controlled models trained in \methodName{}: strong in-distribution performance, sharp degradation under task, length, and format shifts, and limited benefit from increased model capacity. This alignment between synthetic and real-world settings supports the claim that the effectiveness and brittleness of CoT reasoning are fundamentally governed by distributional alignment, rather than by pretraining scale or architectural advances alone.

\section{Additional Qualitative Analysis}
\label{app:qualitative}
To complement the aggregate metrics, we present representative failure traces for each generalization dimension. Each case pairs the model's generated chain with the ground-truth chain and highlights the specific failure mode --- pattern replication for task shifts, length rigidity for length shifts, and template reliance for format shifts.

\subsection{Failures in Task Generalization}
\label{app:qualitative:task_generalization}
Recall from Section~\ref{sec:task_generalization} that task generalization decomposes into transformation and element generalization. We illustrate each with a concrete POOD/OOD failure below.

\subsubsection{Failures caused by Transformation Generalization}
The following case illustrates the failure caused by the transformation partial OOD setting discussed in Section~\ref{app:illustration_of_generalization:transformation}. Specifically, the model is trained exclusively on the compositional transformation $f_1 \circ f_2$, but is evaluated on the unseen composition $f_1 \circ f_1$.
\begin{tcolorbox}[promptstyle]
\ttfamily
Query: \\
H U S P [F1] [F1] <think>\\
Response: \\
U H F C [F2] <answer> H F C U\\
Ground-Truth: \\
U H F C [F1] <answer> H U S P
\end{tcolorbox}
The results demonstrate that the model persists in applying the training composition ($f_1 \circ f_2$) rather than adapting to the new sequence ($f_1 \circ f_1$), even though the constituent transformation $f_1$ was observed during training. The model incorrectly generates the token \texttt{[F2]} in the reasoning step, mirroring the fixed pattern learned from the training data.
\subsubsection{Failures caused by Element Generalization}
The following case illustrates the failure caused by the element-level OOD setting discussed in Section~\ref{app:illustration_of_generalization:element}. Specifically, the training data was restricted to visible elements A--M, while the model was evaluated on queries consisting exclusively of elements N--Z.
\begin{tcolorbox}[promptstyle]
\ttfamily
Query: \\
N N N Q [F2] [F1] <think>\\
Response: \\
A A C E [F1] <answer> N N P R\\
Ground-Truth: \\
N N Q N [F1] <answer> A A D A
\end{tcolorbox}
In the generated reasoning, the model incorrectly outputs \texttt{AACE} instead of the expected intermediate result \texttt{NNQN}. This error arises because all training queries consisted solely of elements A--M. Consequently, the model fails to generalize the transformation $f_2$ to the unseen elements; instead, it reverts to the training distribution, attempting to replicate the A--M patterns observed during training.
\subsection{Failures in Length Generalization}
\subsubsection{Failures in Text Length Generalization}
The following failure case demonstrates the model's inability to generalize to unseen text lengths. In this experiment, the model was trained on sequences of length four but evaluated on sequences of length five.
\begin{tcolorbox}[promptstyle]
\ttfamily
Query: \\
I G L L Q [F1] [F2] <think>\\
Response: \\
T Y Y [F2] <answer> T Y Y V\\
Ground-Truth: \\
V T Y Y D [F2] <answer> T Y Y D V
\end{tcolorbox}
Despite the increased length of the input query (five atoms), the model fails to adapt. As seen in the generated answer (\texttt{T Y Y V}), the model rigidly adheres to the length constraint observed during training, outputting a sequence of four atoms instead of the required five.
\subsubsection{Failures in Reasoning Step Generalization}
\label{app:text_length_example}
The following case shows that a model trained under $f_1 \circ f_1$ tries to reproduce the length in training data by adding tokens in the reasoning chain even when prompted with the seen transformation $f_1$.
\begin{tcolorbox}[promptstyle]
\ttfamily
Query: \\
A A B D [F1] <answer>\\
Response: \\
N O A Z N N O Q [F1] <answer> A A B D\\
Ground-Truth: \\
N N O Q
\end{tcolorbox}
The ground-truth chain has one reasoning step applying $f_1$ once to \texttt{AABD} to obtain \texttt{NNOQ}. Instead, the model pads the reasoning trace to the two-step length it was trained on: it emits an extra eight-atom segment \texttt{N O A Z N N O Q} before the \texttt{[F1]} token, and only then produces \texttt{A A B D}. This is the same pattern observed in aggregate in Fig.~\ref{fig:reasoning_step}: the model treats the number-of-steps statistic of the training distribution as a hard template rather than a flexible property of the input, and reconstructs a two-step-long trace even when a single step suffices.

\subsection{Failures in Format Generalization}
The following failure case demonstrates the model's inability to generalize under format change. Specifically, here we use \emph{delete} as the format-changing mechanism: one of the two transformation tokens that training queries contain (e.g., \texttt{[F1] [F2]}) is removed, so the test query carries only a single \texttt{[F1]} after the element \texttt{A A A T}.
\begin{tcolorbox}[promptstyle]
\ttfamily
Query: \\
A A A T [F1] <answer>\\
Response: \\
N G N G Y\\
Ground-Truth: \\
N N N G
\end{tcolorbox}
Applying $f_1$ once to \texttt{AAAT} should yield \texttt{NNNG}. Instead, the model emits the five-atom string \texttt{N G N G Y}: it retains the F1-rotated alphabet (\texttt{A}{\textrightarrow}\texttt{N}, \texttt{T}{\textrightarrow}\texttt{G}) but pads the answer to the two-step output length the model was trained to produce, and fills the extra positions with near-random atoms drawn from the same rotated vocabulary. The failure is not a clean ``the model didn't recognize the format'' but a more subtle \emph{joint} mishandling of format and length: once the prompt drifts outside the seen template, the model falls back on whatever training-time statistic is most salient (here, answer length and per-position token distribution), producing output that is locally plausible yet globally incorrect. The same pattern appears under the Insert and Modify perturbations in Figure~\ref{fig:format_generalization}: format perturbations do not merely add surface noise but shift the prompt out of the learned template distribution, triggering the same replication-over-generalization behavior observed for task and length shifts.

\section{Experiment Environment and Implementation Details}
\label{app:exp}
This section collects the practical configuration behind both the controlled experiments (LLMs trained from scratch) and the real-world fine-tuning experiments on SOTA LLMs, enabling full reproduction of our results.

\subsection{Environment Setup}
We conduct controlled experiments using LLMs with different sizes (ranging from 62K to 3B) and different architectures (GPT and LLaMA) of LLMs; detailed hyperparameters are summarized in Table~\ref{tab:model_arch_setting}. For LLMs trained from scratch, we employ the AdamW optimizer in mixed precision (FP16). The default learning rate is $3\times10^{-3}$, and the schedule follows a cosine decay with a 10\% warm-up ratio. Training is conducted using a batch size of 1024, and each model is optimized for 10 epochs. A weight decay of 0.01 is applied, and gradient norms are clipped at 1.0. During the inference time, we set the temperature to 1e-5.

For fine-tuning state-of-the-art LLMs, we use a per-device batch size of 16 with 8 gradient accumulation steps (effective batch size of 128) and train for 24K optimization steps. We set the learning rate to 1e-4 and adopt a cosine learning-rate schedule with a 10\% warm-up ratio. All fine-tuning experiments are conducted with bfloat16 (bf16) mixed-precision training. Given the scale of the data, the results of controlled experiments are averaged over three independent runs.

\begin{table*}[!ht]
\centering
\caption{Hyperparameter settings for LLMs with different model sizes and architectures.}
\label{tab:model_arch_setting}
\resizebox{0.7\linewidth}{!}{
\begin{tabular}{l|c|c|c|c|c}
\toprule
\textbf{Architecture} & \textbf{\# Params} & \textbf{Hidden Size} & \textbf{Intermediate Size} & \textbf{\# Layers} & \textbf{\# Heads} \\
\midrule
\multirow{6}{*}{GPT}
 & 68K  & 32   & N/A  & 4  & 4  \\
 & 589K & 80   & N/A  & 7  & 8  \\
 & 4.8M & 256  & N/A  & 6  & 4  \\
 & 35M  & 512  & N/A  & 11 & 8  \\
 & 540M & 1536 & N/A  & 19 & 24 \\
 & 3B   & 3072 & N/A  & 26 & 32 \\
\midrule
\multirow{6}{*}{LLaMA}
 & 62K  & 48   & 128  & 2  & 4  \\
 & 631K & 80   & 216  & 8  & 4  \\
 & 6M   & 288  & 768  & 6  & 6  \\
 & 60M  & 640  & 1728 & 12 & 10 \\
 & 623M & 1536 & 4096 & 22 & 12 \\
 & 3B   & 3072 & 8192 & 26 & 24 \\
\bottomrule
\end{tabular}
}
\end{table*}

\subsection{Computational Cost}
To accelerate our research, we perform training, fine-tuning, and inference using 8 NVIDIA A100 GPUs (80 GB memory each) and 4 NVIDIA H200 GPUs. Toy experiments can be run on a single A100 GPU.

\section{Discussion and Implication}
\label{app:discussion}
Our investigation, conducted through the controlled environment of \methodName{}, reveals that the apparent reasoning prowess of CoT is largely a brittle mirage. The findings across task, length, and format generalization experiments converge on a conclusion: CoT is not a mechanism for genuine logical inference but rather a sophisticated form of structured pattern matching, fundamentally bounded by the data distribution seen during training. When pushed even slightly beyond this distribution, its performance degrades significantly, exposing the superficial nature of the ``reasoning'' it produces.

While our experiments utilized models trained from scratch in a controlled environment, the principles uncovered are extensible to large-scale pre-trained models. We summarize the implications for practitioners as follows.

\textbf{Guard against over-reliance and false confidence.} CoT should not be treated as a ``plug-and-play'' module for robust reasoning, especially in high-stakes domains like medicine, finance, or legal analysis. The ability of LLMs to produce ``fluent nonsense''---plausible but logically flawed reasoning chains---can be more deceptive and damaging than an outright incorrect answer, as it projects a false aura of dependability. Sufficient auditing from domain experts is indispensable.

\textbf{Prioritize OOD testing.} Standard validation practices, where the test set closely mirrors the training set, are insufficient to gauge the true robustness of a CoT-enabled system. Practitioners must implement rigorous \textbf{adversarial and OOD testing} that systematically probes for vulnerabilities across task, length, and format variations. 

\textbf{Recognize fine-tuning as a patch, not a panacea.} Our results show that Supervised Fine-Tuning (SFT) can quickly ``patch'' a model's performance on a new, specific data distribution. However, this should not be mistaken for achieving true generalization. It simply expands the model's ``in-distribution'' bubble slightly. Relying on SFT to fix every OOD failure is an unsustainable and reactive strategy that fails to address the core issue: the model's lack of abstract reasoning capability.

\section{Use of Generative AI}
To enhance clarity and readability, we utilized the GPT-5.2 model exclusively as a language polishing tool. Its role was confined to proofreading, grammatical correction, and stylistic refinement---functions analogous to those provided by traditional grammar checkers and dictionaries. This tool did not contribute to the generation of new scientific content or ideas, and its usage is consistent with standard practices for manuscript preparation.

\end{document}